%% file: spacegan.tex
\documentclass{article}
\usepackage{soul}
\usepackage[preprint,nonatbib]{neurips_2019}
\usepackage[utf8]{inputenc}
\usepackage[T1]{fontenc}
\usepackage{hyperref}
\usepackage{url}
\usepackage{booktabs}
\usepackage{amsfonts}
\usepackage{nicefrac}
\usepackage{microtype}
\usepackage{algorithm}
\usepackage{wrapfig}
\usepackage{xcolor}
\usepackage[noend]{algpseudocode}
\usepackage{graphicx}
\usepackage{caption}
\usepackage{subcaption}
\usepackage{amsmath}
\usepackage{mathtools}
\usepackage{rotating}

\usepackage[square,numbers]{natbib}
\makeatletter
\g@addto@macro{\UrlBreaks}{%
\do\/\do\a\do\b\do\c\do\d\do\e\do\f%
\do\g\do\h\do\i\do\j\do\k\do\l\do\m%
\do\n\do\o\do\p\do\q\do\r\do\s\do\t%
\do\u\do\v\do\w\do\x\do\y\do\z%
\do\A\do\B\do\C\do\D\do\E\do\F\do\G%
\do\H\do\I\do\J\do\K\do\L\do\M\do\N%
\do\O\do\P\do\Q\do\R\do\S\do\T\do\U%
\do\V\do\W\do\X\do\Y\do\Z}
\makeatother

\title{Augmenting correlation structures in spatial data using deep generative models}

\author{%
  Konstantin Klemmer \thanks{Equal contribution} \\
  The Alan Turing Institute \& \\
  University of Warwick \\
  \texttt{kklemmer@turing.ac.uk} \\
  \And
  Adriano Koshiyama \footnotemark[1] \\
  The Alan Turing Institute \& \\
  University College London \\
  \texttt{akoshiyama@turing.ac.uk} \\
  \And
  Sebastian Flennerhag \\
  The Alan Turing Institute \& \\
  University of Manchester \\
  \texttt{sflennerhag@turing.ac.uk} \\
}

\begin{document}

\maketitle

\begin{abstract}
State-of-the-art deep learning methods have shown a remarkable capacity to model complex data domains, but struggle with geospatial data. In this paper, we introduce \emph{SpaceGAN}, a novel generative model for geospatial domains that learns neighbourhood structures through spatial conditioning. We propose to enhance spatial representation beyond mere spatial coordinates, by conditioning each data point on feature vectors of its spatial neighbours, thus allowing for a more flexible representation of the spatial structure. To overcome issues of training convergence, we employ a metric capturing the loss in local spatial autocorrelation between real and generated data as stopping criterion for \emph{SpaceGAN} parametrization. This way, we ensure that the generator produces synthetic samples faithful to the spatial patterns observed in the input. \emph{SpaceGAN} is successfully applied for data augmentation and outperforms compared to other methods of synthetic spatial data generation. Finally, we propose an ensemble learning framework for the geospatial domain, taking augmented \emph{SpaceGAN} samples as training data for a set of ensemble learners. We empirically show the superiority of this approach over conventional ensemble learning approaches and rivaling spatial data augmentation methods, using synthetic and real-world prediction tasks. Our findings suggest that \emph{SpaceGAN} can be used as a tool for (1) artificially inflating sparse geospatial data and (2) improving generalization of geospatial models.
\end{abstract}

\input{introduction}

\input{method}

\input{experiments}

\input{related_work}

\input{conclusion}

\newpage

\subsubsection*{Acknowledgments}

The authors gratefully acknowledge funding from the UK Engineering and Physical Sciences Research Council, the EPSRC Centre for Doctoral Training in Urban Science (EPSRC grant no. EP/L016400/1); The Alan Turing Institute (EPSRC grant no. EP/N510129/1).

\medskip

\small

\bibliography{spacegan,cganbib}

\newpage

\input{appendix}

\end{document}

%% file: introduction.tex
\section{Introduction}\label{Section1}

The empirical analysis of geospatial patterns has a long tradition, with applications ranging from estimating rainfall patterns \citep{Azimi-Zonooz1989} to predicting housing prices \citep{Basu1998}. Recently, machine learning methods have become increasingly popular for these tasks. Traditional techniques to model spatial dependencies include clustering \citep{Huang2013} or kernel methods like Gaussian Processes (GPs) \citep{Datta2016}. Recent years have seen efforts to scale GP models to high-dimensional data \citep{Gardner2018} and the emergence of convolutional neural networks (CNNs) for learning spatial representations \citep{Shi2015}. But while deep learning methods like CNNs improve upon GP models by enabling non-euclidean, graph-structured data \citep{Henaff2015}, they appear to struggle with long-range spatial dependencies \citep{Linsley2018}. A recent review paper by Reichenstein et al. \citep{Reichstein2019} highlights further problems of deep learning applications with spatial data, setting a research agenda aiming to improve the representation of spatial structures, particularly in deep learning methods.

Furthering this agenda, we explore how generative adverserial nets (GANs) \citep{Goodfellow2014} can capture spatially dependent data and how we can leverage them to learn observed spatial patterns. As they preform well on visual data, in the geospatial context GANs have been used for generating satellite imagery \citep{Lin2017}. However, geospatial point patterns---data points distributed across continuous or discrete $2$-dimensional space with one or more feature dimensions---remain unexplored in that regard. While previous studies have examined GAN performance in the presence of one-dimensional autocorrelation, such as temporal point processes \citep{Xiao2017} or financial time-series \citep{Koshiyama2019}, the multi-dimensional correlation structures in geospatial point patterns pose a more complex challenge. We tackle this issue by introducing \textit{SpaceGAN}: Borrowing well established techniques from geographic information science, we use spatial neighbourhoods as context to train a conditional GAN (cGAN) and optimize cGAN selection for the best representation of the inputs local spatial autocorrelation structures. GANs are difficult to train, often failing to converge to a stable solution. Our novel stopping criterion explicitly measures the quality of the representation of observed spatial patterns. Furthermore, this approach enables us to work with data distributed in discrete and continuous space. Respresentations learned by \emph{SpaceGAN} can be used for downstream tasks, even on out-of-sample geospatial locations. We show how this can be used for prediction via an ensemble learning framework. We test our approach on synthetic and real-world geospatial prediction tasks and evaluate the results using spatial cross-validation.  

The main contributions of this study are as follows: First, we introduce a novel cGAN approach for geospatial data domains, focusing on capturing spatial dependencies. Second, we introduce a novel ensemble learning method tailored to spatial prediction tasks by utilizing \emph{SpaceGAN} samples as training data for a set of base learners. Across different experimental settings, we show that \textit{SpaceGAN}-generated samples can substantially improve the performance of predictive models. As such, the results also have practical implications: our proposed framework can be used to inflate low-dimensional spatial data. This allows for enhanced model training and reduced bias by compensating for a lack of training data. We thus improve generalization performance, even when compared to existing methods for data augmentation. The remainder of this paper is structured as follows: Section \ref{Section2} introduces the \textit{SpaceGAN} framework and elaborates on the technical details in respect to the cGAN architecture and spatial autocorrelation representation. In Section \ref{Section3}, we evaluate \textit{SpaceGAN} empirically using synthetic and real-world data and comparing it to existing methods for spatial data augmentation and ensemble learning. Section \ref{Section4}  reviews existing literature related to our study.

%% file: method.tex
\section{SpaceGAN}\label{Section2}

\subsection{Spatial Correlation Structures}

\begin{wrapfigure}{r}{0.4\textwidth}
\vskip -0.1in
\centering
\includegraphics[width=0.4\textwidth]{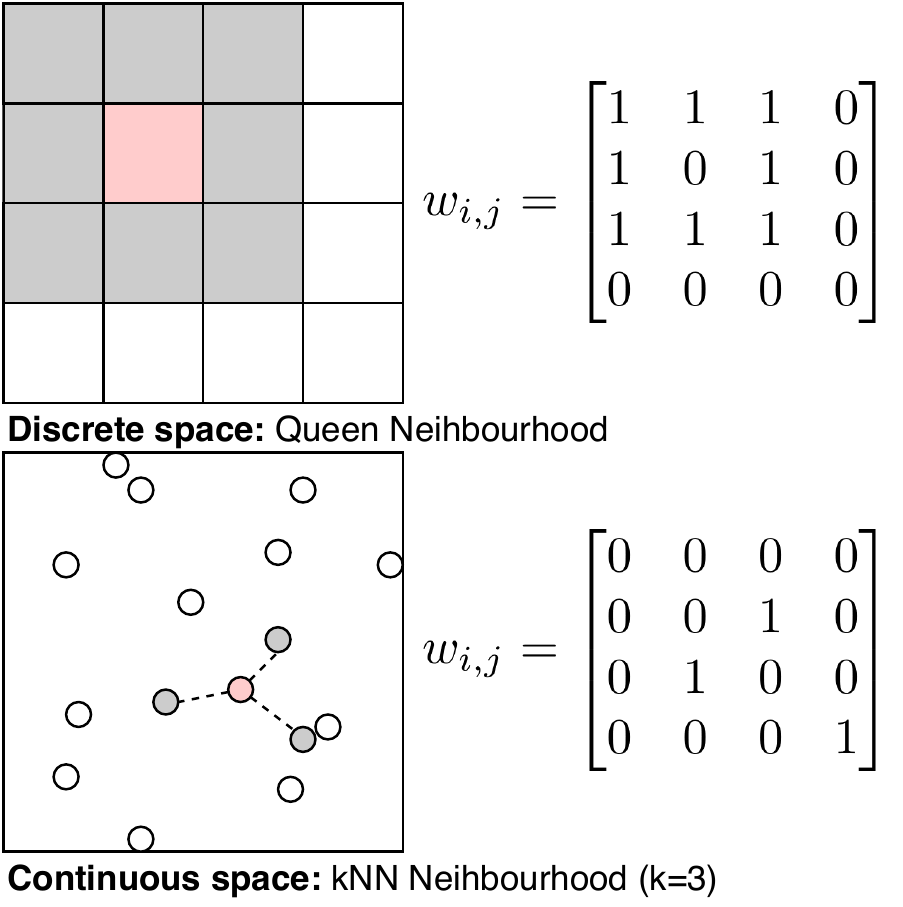}
\vskip -0.1in
\caption{Examples for spatial weight matrices $w_{i,j}$ in the discrete and continuous case.}
\label{fig1}
\vskip -0.15in
%\vskip -0.5in
\end{wrapfigure}
The so-called "First Law of Geography", made famous by Waldo Tobler, states that "everything is related to everything else, but near things are more related than distant things" \citep{Tobler1970}. Following this premise, when working with geospatial data, inherent local inter-dependencies represent an additional information layer that can be exploited. A brief example to illustrate this concept: In a typical city, when we want to estimate the price of a house, we might want to check house prices at nearby locations. If, for instance, the house is located in a rich, spatially contained neighbourhood, just knowing the price of a nearby property and without any further knowledge about the features of the house (e.g. size, age), can provide us with an informed guess. Let us formulate this intuition by first defining the $i$-th data point as a tuple $\mathbf{d}_i = (\mathbf{x}_i, y_i, \mathbf{c}_i)$, where $[x_{i}^{(1)}, ..., x_{i}^{(m)}] = \mathbf{x}_i \in \mathbb{R}^{m}$ describes a set of $m$ features, $y_i \in \mathbb{R}$ describes the target vector and $[c_i^{(1)}, c_i^{(2)}] = \mathbf{c}_i \in \mathbb{R}^{2}$ describes the point coordinates in $2d$ space. While a supervised learning setting with target $y_i$ is not needed, we introduce it here for simplicity since we apply this during the experiments in Section \ref{Section3}. The features $(\mathbf{x}, y)$ can be distributed across space randomly, or follow a---global or local---spatial process. This can be examined by measuring the correlation of a feature with its local neighbourhood, the so called local spatial autocorrelation, which is given by the Moran's I metric \citep{MORAN1950}. While originally theorized for phenomena distributed in $n$-dimensional space, the concept was widely popularized in geostatistics by Luc Anselin \citep{Anselin1995}. His formalization gives a local autocorrelation coefficient for a vector distributed across space. While this can be applied to any vector in the feature set of $\mathbf{d}$, we will explain the concept using the target vector $y$ here. We assume $y$ to follow some spatial process $y \sim f(\mathbf{c})$. As such, $y$ consists of $n$ real-valued observations $y_{i}$ referenced by an index set $N = \{1,2,...,n \}$ indicating the spatial unit corresponding to the coordinate $\mathbf{c}_i$. Let the neighbourhood of the spatial unit $i$ be $\mathcal{N}_{i} = \{ j \in N | \exists i \ni N : w_{i,j} \neq 0 \}$. In accordance with our conceptualization above, we can then compute its local spatial autocorrelation $I_{i} = I(y_{i})$ as:

\begin{equation}
    I_{i} = (n-1) \frac{y_{i} - \bar{y}}{\sum^{n}_{j=1, j \neq i} (y_{j} - \bar{y})^2} \sum^{n}_{j=1, j \neq i} w_{i,j} (y_{j} - \bar{y})
\end{equation}

where $\bar{y}$ represents the mean of $y_{i}$'s and $w_{i,j}$ are components of a weight matrix indicating membership of the local neighbourhood set between the observations $i$ and $j$. For $y_{i}$ distributed in continuous space, the weight matrix can, for example, correspond to a $k$-nearest-neighborhood with $w_{i,j} = 1$ if $j \in \mathcal{N}_{i}$ and $w_{i,j} = 0$ otherwise. For $y_{i}$ distributed in discrete space (e.g. non-overlapping, bordering polygons), the weight matrix could for example correspond to a queen neighbourhood (see Figure \ref{fig1}). The Moran's I metric hence takes in a vector distributed in space and its corresponding neighbourhood structure to calculate how strongly (positively or negatively) the vector is autocorrelated with its spatial neighbourhood at any given location. Intuitively, this makes the selection of the weight matrix $w_{i,j}$, i.e. the definition of "neighbourhood", an important design choice which we have to account for when trying to augment spatial data imitating the spatial autocorrelation structures of the input. For this augmentation process, we turn towards a popular family of generative models: GANs. 

\subsection{Spatially-conditioned GANs}

GANs are a class of models employing two Neural Networks: a Generator ($G$) and a Discriminator ($D$). The Generator is responsible for producing a latent representations of the input, attempting to replicate a given data generation process. It is defined as a neural network $G(\mathbf{z},\Theta_{G})$ with parameters $\Theta_{G}$, mapping noise $\mathbf{z} \sim p_{z}(z)$ to some feature space $\mathbf{x}$ ($G: \mathbf{z} \rightarrow \mathbf{x}$). The Discriminator, a neural network $D(\mathbf{x},\Theta_{D})$, aims to probabilistically distinguish the synthetic input $\mathbf{\hat{x}}$ created by the Generator and real data $\mathbf{x} \sim p_{data}(\mathbf{x})$ ($D:\mathbf{x} \rightarrow [0,1]$). Both networks compete in a minimax game, improving their performance until the real and synthetic data are undistinguishable from one another. But while GANs have been successfully applied in many areas, training them is highly non-trivial\cite{salimans2016improved,gulrajani2017improved} and remains an area of intense study \cite{arjovsky2017wasserstein,gulrajani2017improved,wang2018evolutionary,mao2017least}. This is further complicated by the non-$iid$ nature of geospatial data, in which learning an unconditional model would ignore inherent local dependencies. To overcome this, a sampling process taking spatial structure into account is needed, thus preserving statistical properties such as local spatial autocorrelation. 

Therefore, conditional GANs (cGANs) \cite{mirza2014conditional} are better fit to handle context-dependent data generation, such as geospatial data. In cGANs, the input to both the generator and discriminator are augmented by a context vector $\mathbf{v}$. Typically, $\mathbf{v}$ represents a class label that we want the cGAN to generate an input for, but it can be any form of contexualization. Formally, we can define a cGAN by including the conditional variable $\mathbf{v}$ in the original formulation so that $G: \mathbf{z} \times \mathbf{v} \rightarrow \mathbf{x}$ and $D: \mathbf{x} \times \mathbf{v} \rightarrow [0, 1]$. The minimax game between $D$ and $G$ is then given as $V(G,D)$:

\begin{equation}
\min_G \max_D V(D,G) = \mathbb{E}_{\mathbf{x} \sim p_{data}(\mathbf{x})} \bigl[\log D(\mathbf{x} | \mathbf{v})\bigr] + \mathbb{E}_{\mathbf{z} \sim p_{\mathbf{z}}(\mathbf{z})} \bigl[\log (1 - D(G(\mathbf{z} | \mathbf{v})))\bigr]
\end{equation}

cGANs have previously been used for spatial conditioning of image data, using pixel coordinates. In our formulation, this would translate to setting $\mathbf{v} = \mathbf{c}$ \citep{Lin2019,Hu2017}. However, this approach is not sufficient for our problem since mere conditioning on the point coordinate alone would omit valuable information about the local neighbourhood of each point. Instead, for each point $\mathbf{d}_{i}$ we are interested in capturing how its features $(\mathbf{x}_{i},y_{i})$ relate to those of neighbouring points $(\mathbf{x}_{j},y_{j}) \in \mathcal{N}_{i}$. As such, we define the \emph{SpaceGAN} context vector $\mathbf{v}$ of point $i$ as $\mathbf{v} = \mathcal{N}_{i}$. 

Similarly to our intuition of spatial autocorrelation, outlined above, we assume that the features of nearby data points may offer valuable information on the point-of-interest. By conditioning each data point on all neighbouring points we allow for the learning of local patterns across the feature space. Beyond this, the versatility of constructing spatial weights $w_{i,j}$ enables experimentation with and optimization of different spatial neighbourhood definitions. This offers a flexibility that is not provided by point coordinate conditioning.

%\noindent in our case, given a set of spatially referenced features (e.g., n nearest neighbours) \seb{This is new notation, what does $x_1(c_1)$ mean?} $x_{1} (c_1), ..., x_{i} (c_i), ..., x_{k} (c_k)$, our conditional set is $\mathbf{v} = \bigl(x_{1} (c_1), ..., x_{i} (c_i), ..., x_{k} (c_k)\bigr)$ and what we are aiming to sample/discriminate is \seb{confusing: above x is the feature vector and y the target variable, now they are the same thing? Or are they different?} $\mathbf{x} = y_{i} (c_i)$. In this sense, $k$ sets the amount of past\seb{is there a temporal dimension involved? If not, just say number of neighbours} neighbours that is considered in the implicit conditional generative model. If $k = 0$, then a traditional GAN will be trained; if $k$ is large, than the Neural Network have a larger memory\seb{Only true if there's a sequential process to generation. Is there?}, but it will need bigger capacity to model and deal with selecting the right neighbours and dealing with noise vector $\mathbf{z}$.

\subsection{Training and Selecting Generators for Spatial Data}

One problem concerning GANs is that they typically fail to converge to a stable solution. To overcome this, we seek to tie training convergence to some measure of quality of the synthesized data. Accordingly, we propose to evaluate the generator performance by the faithfulness of its produced spatial patterns in relation to the true patterns observed in the input. For this, we introduce a new metric, the Mean Moran's I Error (MIE). It is defined as the mean absolute difference between the local spatial autocorrelation of the input $I(y_1, ..., y_n)$ versus that of the generated samples $I(\hat{y}_1, ..., \hat{y}_n)$:
\begin{equation}
    MIE = \sum^{n}_{i=1} |(I(y_{i}) - I(\hat{y_{i}}) )|
\end{equation}

We apply this metric for model selection by choosing the model that minimizes $MIE$, i.e. the loss of local spatial autocorrelation between real and generated $\hat{y}_1, ..., \hat{y}_n$. In our supervised learning setting, we are particularly interested in a faithful representation of the target vector and hence use $y_1, ..., y_n$ to calculate $MIE$. Of course, $MIE$ can also be calculated using any other feature vector from $\mathbf{d}$. An implementation for multidimensional input is also formalized by Anselin \citep{Anselin2019} or can be achieved by averaging $MIE$ through multiple features. To train \emph{SpaceGAN}, we proceed as when training a normal cGAN, but include the $MIE$ stopping-criterion. Algorithm \ref{training} details our training procedure.

\begin{algorithm*}[h!]
	\caption{SpaceGAN Training and Selection}\label{training}
	\begin{algorithmic}[1]
	    \Require{$snap$, $C$, $L$: hyper-parameter}
		%\Procedure{cGAN}{$[y_1,...,y_T], params$}
		\For{number of training steps ($tsteps$)}
		\State Sample minibatch of $L$ noise samples $\{\mathbf{z}_1, ..., \mathbf{z}_L\}$ from noise prior $p_{\mathbf{z}} (\mathbf{z})$
		\State Sample minibatch of $L$ examples from $p_{data}\bigl((y_{i}, \mathbf{x}_i) \ | \ \mathcal{N}_{i} \bigr)$
		\State Update the discriminator by ascending its stochastic gradient:
		\begin{equation}
		\nabla_{\Theta_D} \frac{1}{L} \sum_{i=1}^{L} \Big[\log D( (y_{i}, \mathbf{x}_i)  \ | \ \mathcal{N}_{i}) + \log (1 - D(G(\mathbf{z}_i \ | \ \mathcal{N}_{i})))\Big] \nonumber
		\end{equation}
		\State Sample minibatch of $L$ noise samples $\{\mathbf{z}_1, ..., \mathbf{z}_L\}$ from noise prior $p_{\mathbf{z}} (\mathbf{z})$
		\State Update the generator by ascending its stochastic gradient:
		\begin{equation}
		\nabla_{\Theta_G} \frac{1}{L} \sum_{l=1}^{L} \Big[\log (D(G(\mathbf{z}_i \ | \ \mathcal{N}_{i})))\Big] \nonumber
		\end{equation}
		\If{$tsteps \ \% \ snap$}
		\State $G_k \gets G$, $D_k \gets D$, $MIE \leftarrow 0$ \Comment{store current $G$, $D$ as $G_k$, $D_k$; initiate $MIE$}
		\For{$C$} \Comment{draw $C$ samples from $G_k$}
		\For{$i\gets 1, n$} \Comment{generate spatial data}
		\State sample noise vector $\mathbf{z} \sim p_{\mathbf{z}} (\mathbf{z})$
		\State draw $(\hat{y}_{i}, \mathbf{\hat{x}}_{i}) = G_k(\mathbf{z} \ | \  \mathcal{N}_{i})$ 
		\EndFor
		\State Measure \emph{SpaceGAN} $y$ samples spatial autocorrelation goodness-of-fit: 
        \begin{equation}
            MIE \leftarrow MIE + \sum^{n}_{i=1} |(I(y_{i}) - I(\hat{y}_{i}) )|
        \end{equation}
		\EndFor
		\State Average of all samples: $MIE(G_k) = \frac{1}{C} MIE$
		\EndIf
		\EndFor
		\State \textbf{return} $G := arg\min_{G_k} MIE(G_k)$, $D := arg\min_{G_k} MIE(G_k)$
		%\EndProcedure
	\end{algorithmic}
\end{algorithm*}

The set of user-defined hyperparameters for running \textit{SpaceGAN Training and Selection} mainly encompass: $G$ and $D$ architectures, number of lags $p$, noise vector size and prior distribution, minibatch size $L$, number of epochs, snapshot frequency ($snap$), number of samples $C$ as well as parameters associated to the stochastic gradient optimizer. For a precise description of the architecture and specific settings, see the experiments in Section \ref{Section3} for details. Notably, our proposed stopping criterion can be seen as choosing the best member from a population of GANs acquired during training. In this way, our approach resembles "snapshot ensembling", introduced by Huang et al. \citep{Huang2017}. 

%We seek to train our conditional GAN as to not only generate believable samples, but also to generate realistic representations of spatial context. We account for this by applying a condition to the GAN selection process: we seek to select the GAN which best imitates the spatial correlations observed in the input. As mentioned above, the selection of the weight matrix $w_{i,j}$, i.e. the underlying definition of "neighbourhood" an important design choice. It could either be chosen according to some prior knowledge about the underlying spatial process, or treated as a hyperparameter in the GAN selection process. For the primer of this study, we chose the neighbourhood weights manually, testing different configurations. However our implementation of the problem can easily be expanded to automatically optimize for the optimal neighbourhood definition.

\subsection{Ganning: GAN augmentation for ensemble learning}

\begin{figure}[!htbp]
%\vskip 0.4in
%\begin{wrapfigure}{r}{0.8\textwidth}
\vskip -0.1in
\begin{center}
\centerline{\includegraphics[scale=0.7]{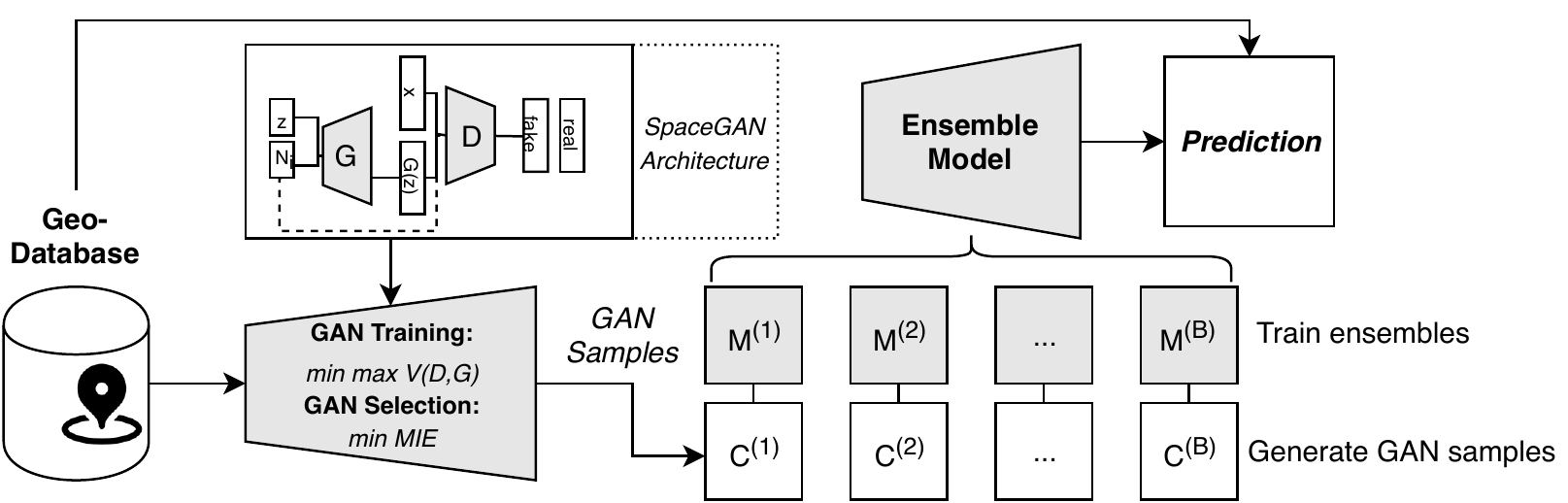}}
\caption{Architecture of \textit{SpaceGAN} for ensemble learning (Ganning).}
\label{fig2}
\end{center}
\vskip -0.2in
%\end{wrapfigure}
\end{figure}

\begin{wrapfigure}{r}{0.6\textwidth}
\vskip -0.2in
\begin{minipage}{0.6\textwidth}
\begin{algorithm}[H]
	\caption{"Ganning" for ensemble learning}\label{cganning}
	\begin{algorithmic}[1]
        \Require{$B$ (number of samples), $M$ (base learner), $G$}
		\For{$b\gets 1, B$} \Comment{generate spatial data}
    		\For{$i\gets 1, n$} 
    		    \State sample noise vector $\mathbf{z} \sim p_{\mathbf{z}} (\mathbf{z})$
    	    	\State draw $(\hat{y}_{i}, \mathbf{\hat{x}}_{i}) = G_k(\mathbf{z} \ | \  \mathcal{N}_{i})$ 
    	    	\EndFor
		\State train base learner: $M^{(b)} (\mathbf{\hat{y}}, \mathbf{x})$
		\EndFor 
		\State \textbf{return ensemble} $M^{(1)}, ..., M^{(B)}$
	\end{algorithmic}
\end{algorithm}
\end{minipage}
\end{wrapfigure}
A common use-case of geospatial data is spatial prediction. We approach this from an ensemble learning perspective. In ensemble learning, individually "weak" base learners (e.g. Regression Trees) can be aggregated and as such outperform "strong" learners (e.g. Support Vector Machines). Traditionally, this idea include models like Random Forest, Gradient Boosting Trees and other implementations that make use of Bagging, Boosting or Stacking principles \cite{friedman2001elements,efron2016computer}. Here, we follow Koshiyama et al. \citep{Koshiyama2019} and utilize \emph{SpaceGAN}-generated samples as training data for the ensemble learners. This approach has not been applied to spatial data before, and since it is analogous to Bagging, we will refer to it as "Ganning" from hereon. Algorithm \ref{cganning} outlines this approach. Assuming a fully trained and parametrized \emph{SpaceGAN}, we repeatedly draw \emph{SpaceGAN} samples and train a base learner for each. After repeating this for $b=1,...,B$ samples we return the whole set of base models $M^{(1)},...,M^{(B)}$ as an ensemble. The benefits of ensemble learning schemes can be best explained using the variance reduction lemma \cite{friedman2001elements}. Intuitively, we can reduce the variance of the ensemble by averaging many weakly correlated predictors. Following the concept of bias-variance trade-off \cite{friedman2001elements,efron2016computer}, the ensemble Mean Squared Error (MSE) decreases, particularly when low bias and high variance base learners such as Deep Decision Trees are used. Nevertheless, there is a potential risk factor to this approach. Should \emph{SpaceGAN} fail to replicate the true data generation process $p_{data}$ truthfully, \emph{SpaceGAN} samples might not only be more diverse, but also more "biased". Consequentially, this could lead to base learners missing obvious patters, or finding new patters that do not exist in the real data. 

%% file: experiments.tex
\section{Experiments}\label{Section3}

%In order to test the applicability of our spatial data augmentation approach, we run experiments on two synthetic and one real world dataset. For all experiments, we first compute the status-quo using just the original training data for model calibration(\textbf{\textcolor{cyan}{Original}}), evaluating on the held-out test set. We compare this case to an artificial inflation of the training set. We first compute three baseline models: \textit{(1)} a spatial bootstrap approach (\textbf{\textcolor{cyan}{SBoot}}), randomly sampling (with replacement) data points from the training set including their immediate spatial neighbourhood. \textit{(2)} A synthetic point generator with naive interpolation (\textbf{\textcolor{cyan}{SP-Naive}}). Here, we create points at new spatial locations and interpolate the values at each point as the mean of its neighbourhood. \textit{(3)} A synthetic point generator with multi-task Gaussian Process (\textbf{\textcolor{cyan}{SP-mGP}}) interpolation, using the same point generation technique as. The original and comparative approaches are then compared to data augmented with \textit{SpaceGAN} at the same synthetic spatial locations (\textbf{\textcolor{cyan}{SP-SpaceGAN}}).

\begin{figure}[!ht]
\centering
\begin{subfigure}{.5\textwidth}
    \includegraphics[width=\linewidth]{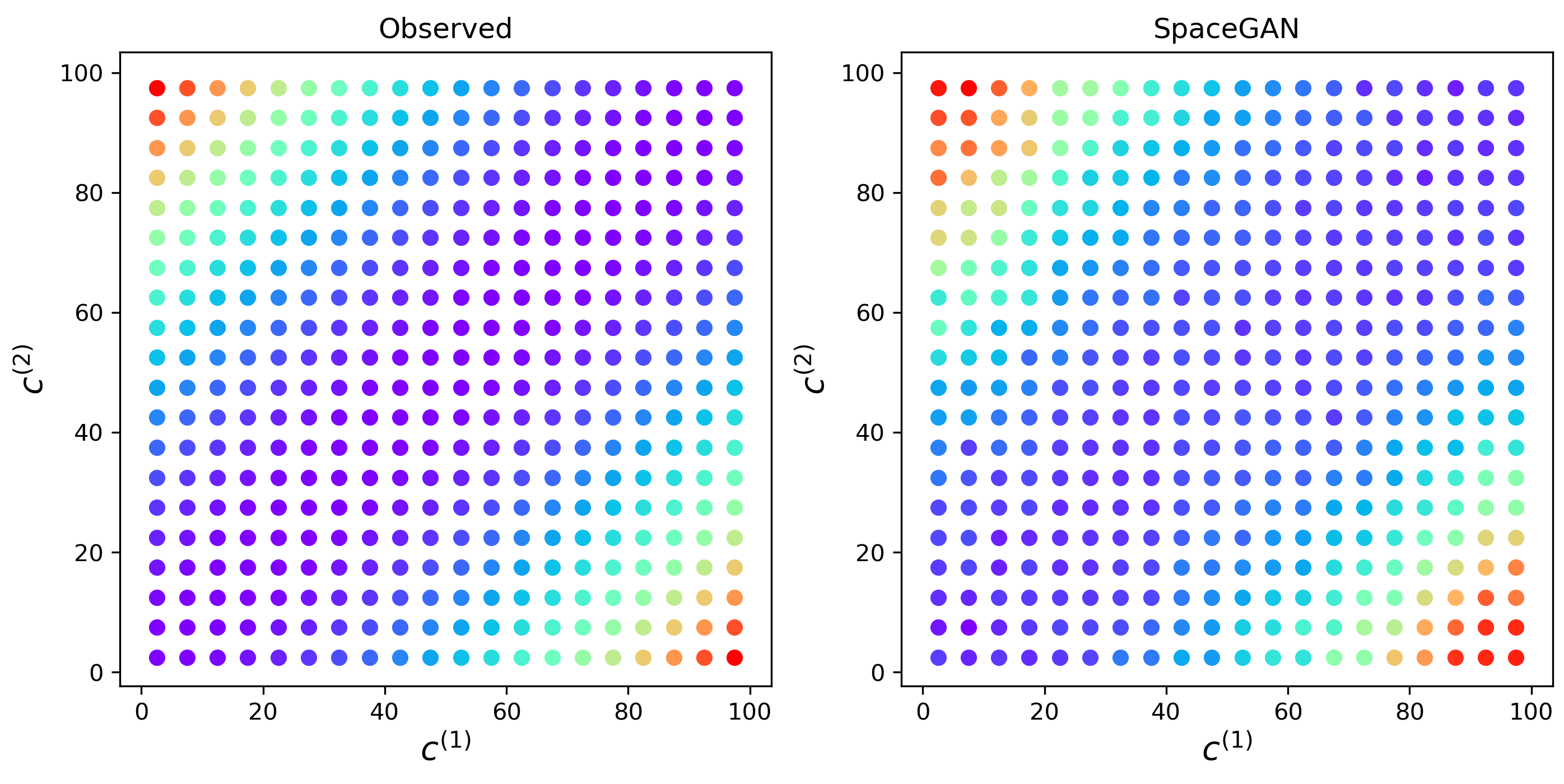}
    \includegraphics[width=\linewidth]{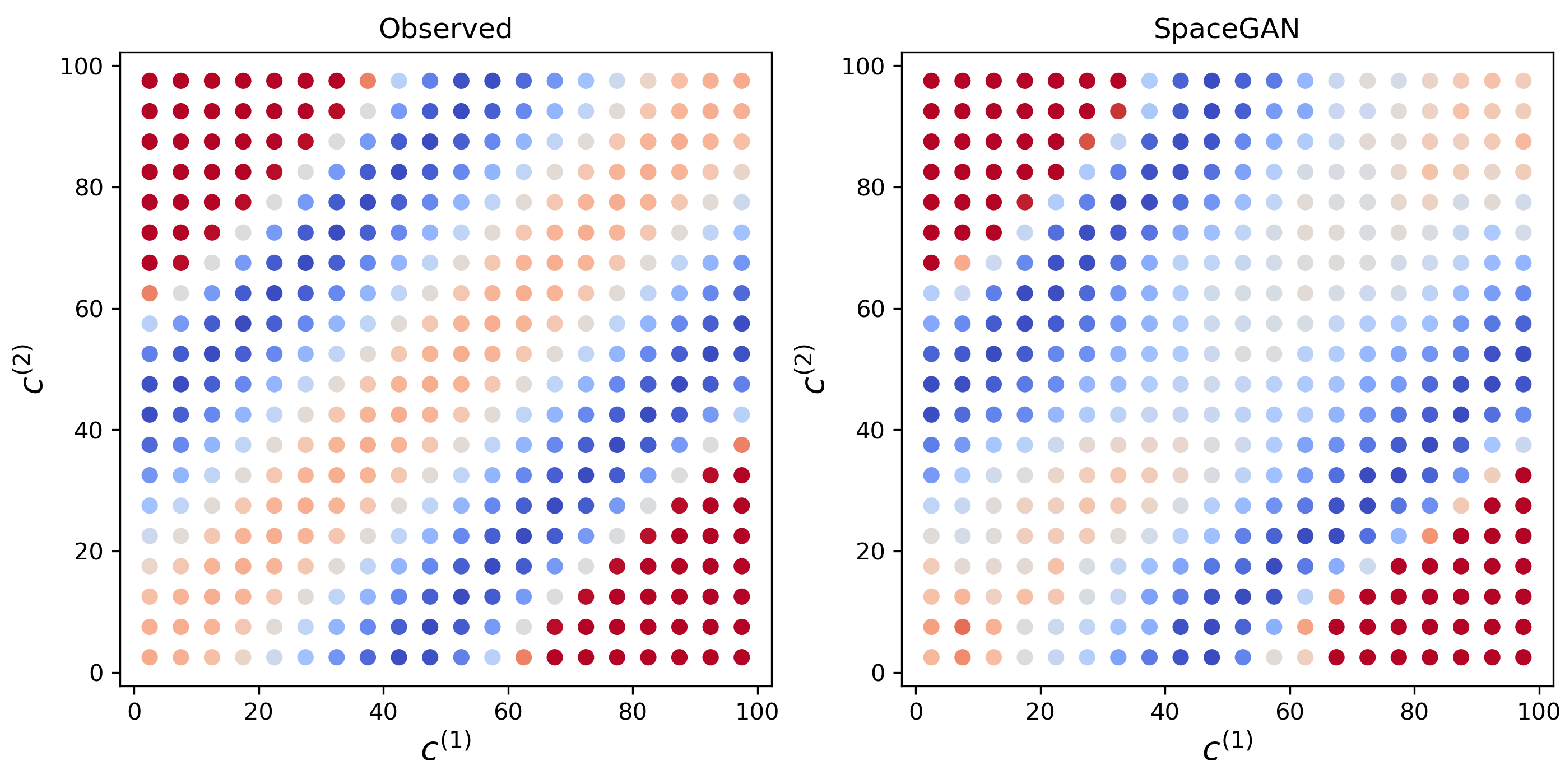}
    \caption{Target vector $y$ (\textit{top}) and its Moran's I value \\ $I(y)$ (\textit{bottom}) of the observed and \emph{SpaceGAN} \\ 
    generated data for \textbf{Toy 1}.}
    \label{fig3:artificial1-part1}
\end{subfigure}%
\begin{subfigure}{.5\textwidth}
    \includegraphics[width=\linewidth]{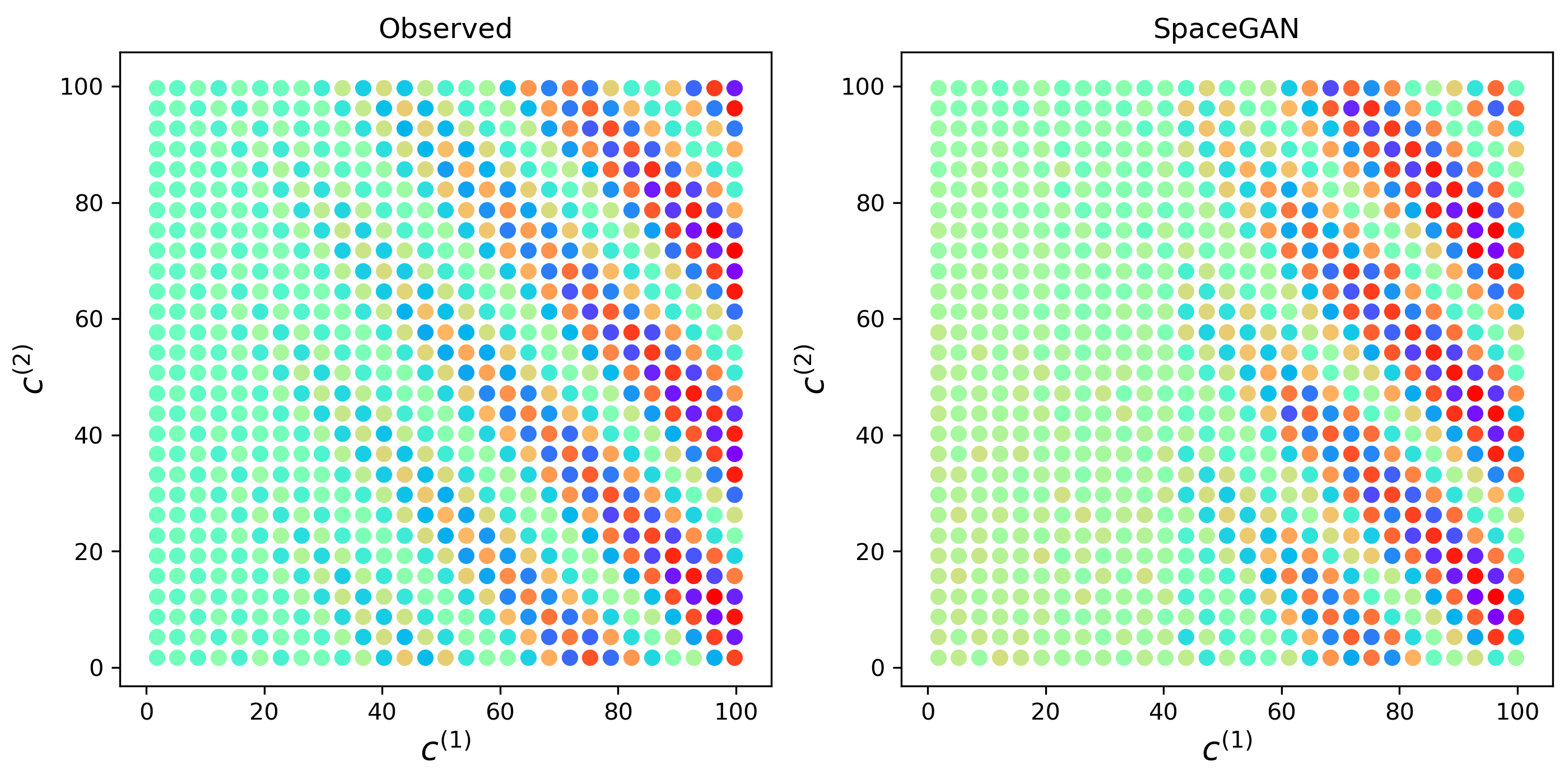}
    \includegraphics[width=\linewidth]{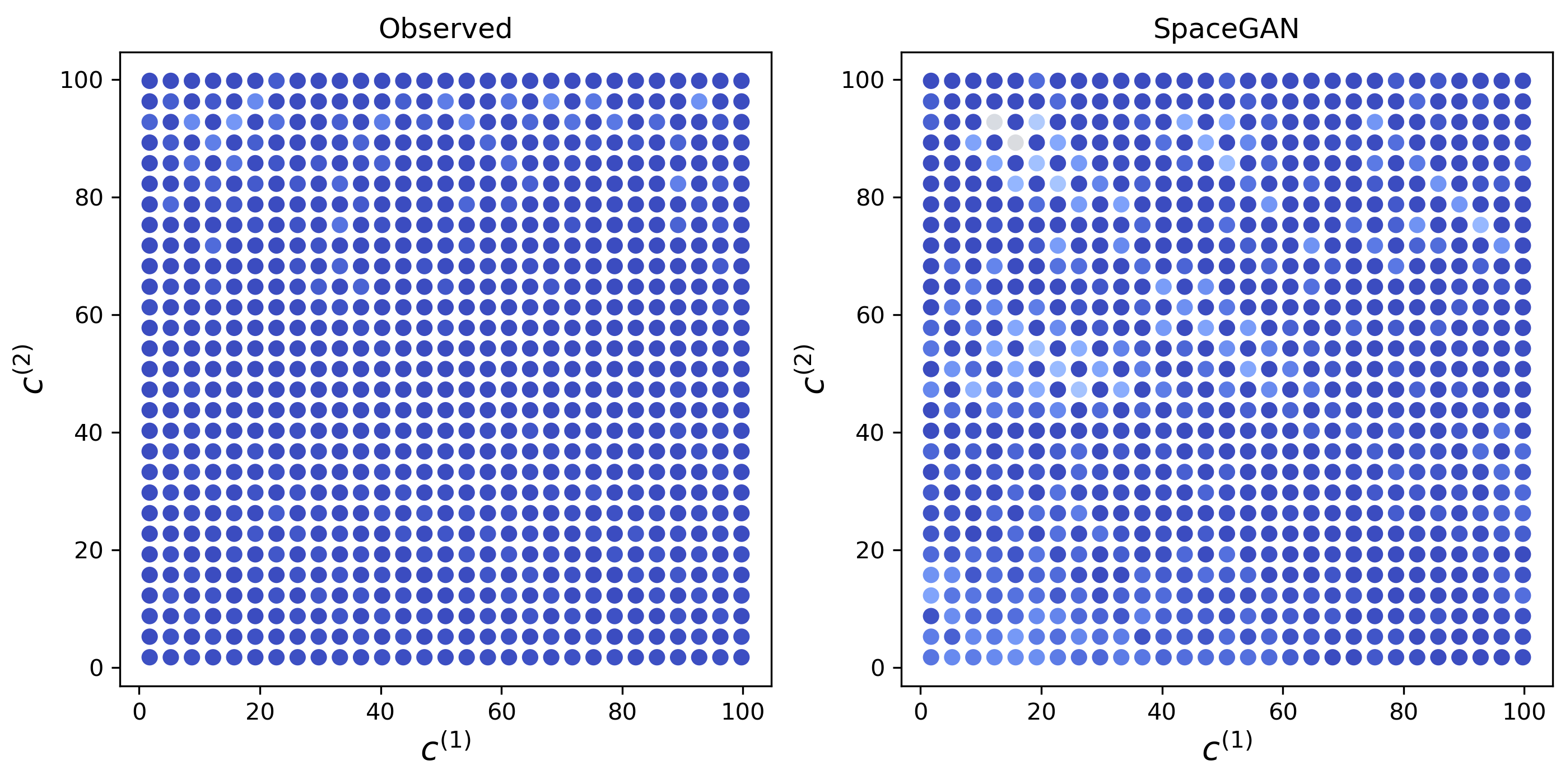}
    \caption{Target vector $y$ (\textit{top}) and its Moran's I value \\
    $I(y)$ (\textit{bottom}) of the observed and \emph{SpaceGAN} \\
    generated data for \textbf{Toy 2}.}
    \label{fig3:artificial2-part1}
\end{subfigure}
\begin{subfigure}{1\textwidth}
    \includegraphics[width=\linewidth]{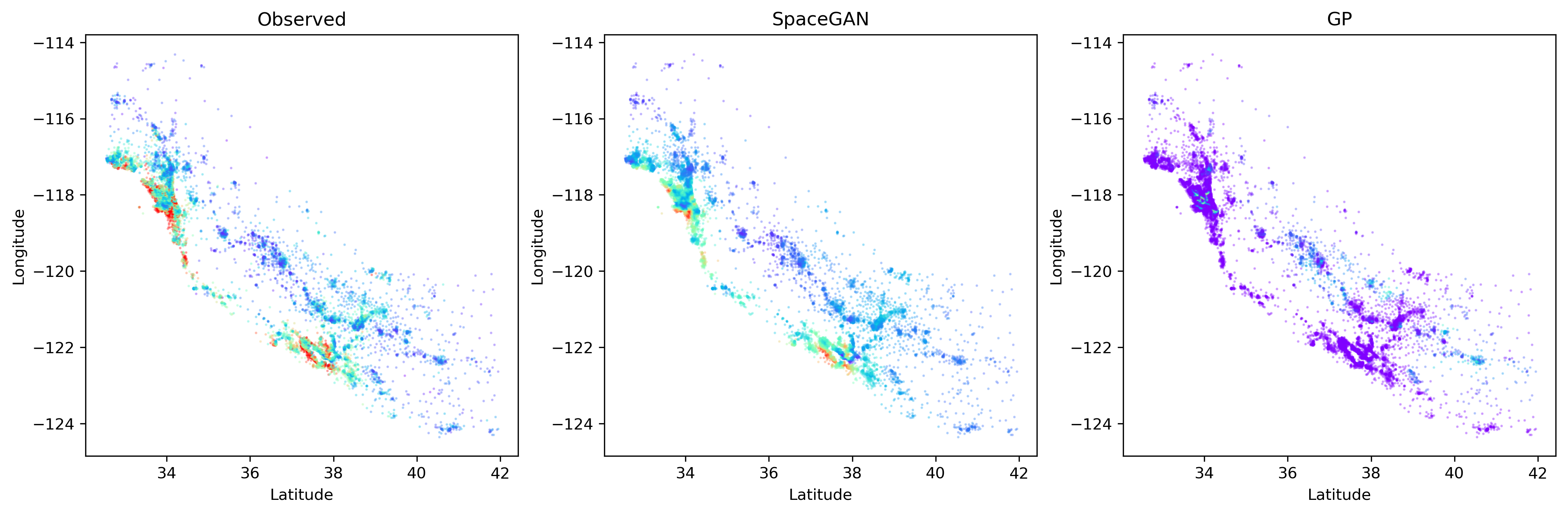}
    \includegraphics[width=\linewidth]{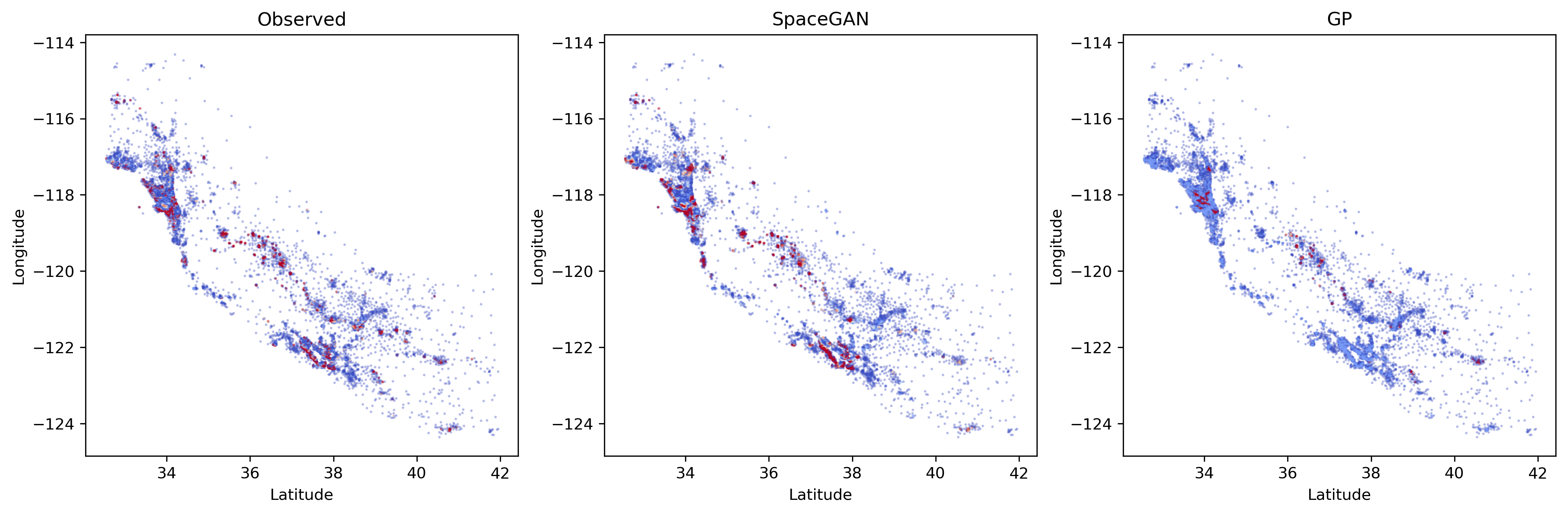}
    \caption{Target vector $y$ (\textit{top}) and its Moran's I value $I(y)$ (\textit{bottom}) of the observed data, \emph{SpaceGAN} generated data and a Gaussian Process smooth for \textbf{California Housing 50}.}
    \label{fig3:housing50-part1}
\end{subfigure}
\caption{\textit{Experiment 1:} We compare the real data to \emph{SpaceGAN} generated samples (averaged over 500 samples) showing both the target $y$ and its Moran's I value $I(y)$. The data is synthesized out-of-sample using spatial cross-validation.} \label{fig3}
\vskip -0.2in
\end{figure}

We evaluate our proposed methods in two experiments. First, we assess \textit{SpaceGAN}'s ability to generate spatial data, including realistic representations of its internal spatial autocorrelation structure. Second, we analyze the use of \textit{SpaceGAN} samples in an ensemble learning approach for spatial predictive modeling. For this, we use three different datasets:

\textbf{Toy 1:} The data points $\mathbf{d}$ are a rectangular grid of $n=400$ regularly distributed, synthetic point coordinates $\mathbf{c}$, a random Gaussian noise vector $\mathbf{x}$ and an outcome variable $y$, a simple quadratic function of the spatial coordinates $\mathbf{c}$ and random vector $\mathbf{x}$.\\
\textbf{Toy 2:} The data points $\mathbf{d}$ are a rectangular grid of $n=841$ regularly distributed, synthetic point coordinates $\mathbf{c}$, a random Gaussian noise vector $\mathbf{x}$ and an outcome variable $y$. Here, $y$ is a more complex combination of a $\pi$-function, a $sin$-function and a linear global pattern of $\mathbf{c}$ and $\mathbf{x}$.\\
\textbf{California Housing:} This real-world dataset describes the prices of $n=20,640$ California houses, taken from the 1990 census. The house prices $y$ come with point coordinates $\mathbf{c}$ and some further predictor variables $\mathbf{x}$, such as house age or number of bedrooms. The dataset was introduced by Pace and Barry \citep{KelleyPace2003} and is a standard example for continuous, spatially autocorrelated data. 

All our experiments are conducted using 10-fold spatial cross-validation \citep{Pohjankukka2017}. Here, points spatially close to the test set are removed from the training set. This is done to prevent overfitting in spatial prediction tasks, as including spatially close and---assuming spatial dependencies---hence similar data to the test set during training can lead to overconfident predictions. For a further elaboration on this scheme, see the Appendix. Note that for the real-world dataset, we refer to \textbf{California Housing 15} as a $15$-nearest neighbour implementation of the spatial cross-validation, and \textbf{California Housing 50} as a $50$-nearest neighbour implementation. For both toy datasets, we use simple queen neighbourhood (see Figure \ref{fig1}). For a description of the specific neural network architectures for \emph{SpaceGAN} used in the different experiments, see Appendix for details. 

\subsection{\textit{Experiment 1:} Reproducing spatial correlation patterns}

\begin{wraptable}{r}{0.6\textwidth}
\vskip -0.1in
\centering
\caption{$MIE$ (and its standard error) between real and augmented data for \emph{SpaceGAN} and GP implementations} \label{table1}
\scalebox{0.8}{
\begin{tabular}{l|cc}
\hline
\hline
& \multicolumn{2}{c}{$MIE$} \\
Dataset & GP$^*$ & SpaceGAN$^*$ \\
\hline
Toy 1 & 1.9495 (0.1750) & 0.3173 (0.1791) \\
Toy 2 & 0.2195 (0.0175) & 0.2141 (0.0157) \\
California Housing 15 & 1.9932 (0.0826) & 1.1468 (0.0416)  \\
California Housing 50 & 3.8183 (0.2072) & 0.9333 (0.0288)  \\
\hline
\hline
\end{tabular}}
\\
$^*$ - output and prediction were normalized before calculation.
\end{wraptable}
Our first experiment aims to investigate \emph{SpaceGANs} ability to not only generate data, but also its capability of reproducing observed spatial patterns. We train \emph{SpaceGAN} on the three experimental datasets and at each spatial location $\mathbf{c}$ return $500$ samples from the generator, as shown in Figure \ref{fig3}. Note that these results show out-of-sample extrapolations. For the dataset \textbf{Toy 1}, \emph{SpaceGAN} is able to capture both the target vector and its spatial autocorrelation almost perfectly. In \textbf{Toy 2}, which represents a substantially more complicated pattern, we capture parts of the observed pattern seamlessly, however the spatial areas characterized by more subtle patterns are not captured fully. Nevertheless, this result shows that \emph{SpaceGAN} also works when the spatial correlation structure is homogeneous. Lastly, we assess the real-world dataset \textbf{California Housing}. Again, \emph{SpaceGAN} is able to capture both the target and the spatial dependencies in the data. In the real-world setting we also compare \emph{SpaceGAN} to a Gaussian Process (GP) smooth for data augmentation (implemented as Vanilla-GP with RBF kernel in sklearn \citep{Pedregosa2012}). We can see that the GP struggles with capturing both, the target vector and its local spatial autocorrelation. Table \ref{table1} provides the $MIE$ metric for \emph{SpaceGAN} and a GP smooth, showing that \emph{SpaceGAN} is best capable of capturing the spatial interdependencies in the input. Higher resolution figures and GP comparisons for \textbf{Toy 1} and \textbf{Toy 2} can be found in the appendix.

\subsection{\textit{Experiment 2:} Data augmentation for predictive modeling}

\begin{figure}[h!]
\begin{center}
    \centerline{\includegraphics[scale=0.35]{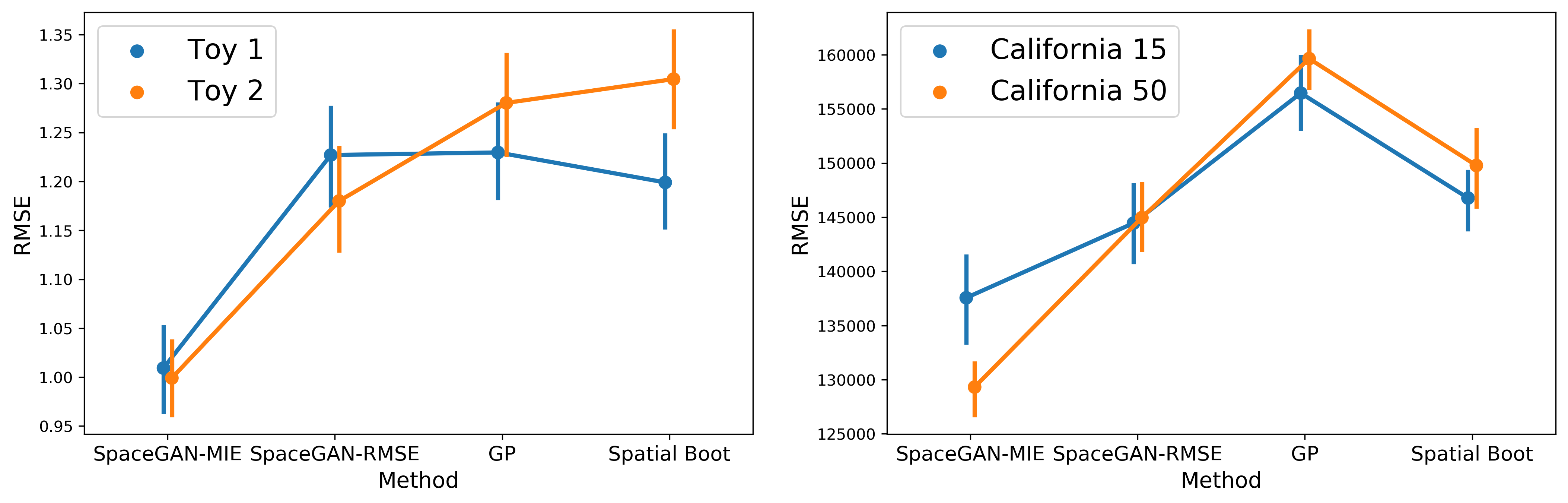}}
    \caption{Model performance of \emph{SpaceGAN} and competing methods across different datasets. $RMSE$ values are given for out-of-sample prediction using spatial cross-validation.}\label{fig4}
\end{center}
\vskip -0.2in
\end{figure}

Our second experiment focuses on predictive modelling in a spatial setting. As outlined in section 2.4, we seek to use \emph{SpaceGAN}-generated samples in an ensemble learning setting---so called "Ganning". More specifically, we test two \emph{SpaceGAN} configurations: First, a \emph{SpaceGAN} using $MIE$ as convergance criterion, second, a \emph{SpaceGAN} using $RMSE$ for convergence. These are compared to two comparable ensemble baselines: First, a GP-Bagging approach, where we draw $B$ samples from a fully trained Gaussian Process posterior and use these to train base models for ensembling (\textbf{GP}). Second, a traditional Bagging approach using spatial bootstrapping (\textbf{Spatial Boot}). Table \ref{table2} provides the out-of-sample prediction $RMSE$ values for the four approaches. Figure \ref{fig4} highlights the average $RMSE$ (with confidence intervals bars) with $B=100$. We can observe that \emph{SpaceGAN} (with $MIE$ convergence) outperforms the competitors by a substantial margin on all three datasets.

% \usepackage{multirow}
\begin{table}[h!]
\centering
\caption{\textit{Experiment 2:} Prediction scores ($RMSE$) and their standard errors across $10$ folds for different ensemble methods with $100$ samples across the different prediction tasks.}
\label{table2}
\scalebox{0.8}{
\begin{tabular}{l|cccc}
\hline
\hline
 & \multicolumn{4}{c}{Model (B = 100)}  \\
Dataset & SpaceGAN-MIE & SpaceGAN-RMSE & GP & Spatial Boot        \\
\hline
Toy 1 & 0.9921 (0.0995) & 1.1993 (0.1494) & 1.2388 (0.1490) & 1.2013 (0.1366) \\
Toy 2 & 1.0097 (0.1092) & 1.2065 (0.1496) & 1.3135 (0.1443) & 1.2962 (0.1413) \\
California Housing 15 & 139534 (12026) & 143983 (10341)  & 159340 (8550) & 148830 (8660)              \\
California Housing 50 & \textbf{128756} (7463) & \textbf{145612} (7152) & \textbf{156814} (8718) & \textbf{148546} (8611) \\
\hline
\hline
\end{tabular}}
\end{table}

%% file: related_work.tex
\section{Related work}\label{Section4}

We now want to contextualize our findings in relation to existing work in the field. As the academic field of machine learning advances, more and more sophisticated techniques are being developed with the aim to capture the complexity of the real world they are trying to model. This is particularly true for spatial methods, where assumptions like distributive independence or Euclidean distances restrict the performance of the most common algorithms. The motivation for this study originates from recent approaches of a more explicit modeling of spatial context within machine learning techniques. Among these are the emergence of vector embeddings for spatially distributed image data \citep{Jean2019}, the opportunities to model non-Euclidean spatial graphs using graph convolutional networks (GCNs) \cite{Defferrard2016} and the modelling of spatial point processes using matrix factorization \citep{Miller2014}. We see \textit{SpaceGAN} as an addition to the family of spatially explicit machine learning methods.

GAN models already have been applied to data autocorrelated in one dimensional space, e.g time series \citep{Xiao2017,Koshiyama2019}, two dimensional space, e.g. remote sensing imagery \citep{Lin2017,Zhu2019} and even three dimensional space, e.g. point clouds \citep{Li2018,Fan2017}. However, none of this previous work used measures of local autocorrelation to improve the representation of spatial patterns. 
In the context of data augmentation, GANs have become a popular tool for inflating training data and increasing model robustness \citep{Xiao2017,Taylor2019,Frid-Adar2018,Bowles2018}. However, such a method does not exist yet for $2d$ multivariate point data, where techniques such as the spatial bootstrap \citep{Brenning2012} or synthetic point generators \citep{Li2016,Quick2015} are most commonly used. Spatial image data and point clouds on the other hand are often augmented using random perturbations, rotations or cropping \citep{Gerke2016,Zhou2018}.
Lastly, ensemble learning is increasingly popular for spatial modeling \citep{Davies2016}, with applications ranging from forest fire susceptibility prediction \citep{Tehrany2018} to class ambiguity correction in spatial classifiers \citep{Jiang2017}. Nevertheless, to our knowledge, no research has yet been conducted combining GAN augmentation and ensemble learning within a spatial data environment, highlighting the novelty of this study.

%% file: conclusion.tex
\section{Conclusion}\label{Section5}

In this paper we introduce \emph{SpaceGAN}, a novel data augmentation method for spatial data, reproducing both the data structure and its spatial dependencies through two key innovations: First, we provide a novel approach to spatially condition the GAN. Instead of conditioning on raw spatial features like coordinates, we use the feature vectors of spatially near data points for conditioning. Second, we introduce a novel convergence criterion for GAN training, the $MIE$. This metric measures how well the generator is able to imitate the observed spatial patterns. We show that this architecture succeeds at generating faithful samples in experiments using synthetic and real-world data. Turning towards predictive modeling, we propose an ensemble learning approach for spatial prediction tasks utilizing augmented \emph{SpaceGAN} samples as training data for an ensemble of base models. We show that this approach outperforms existing methods in ensemble learning and spatial data augmentation. 

In developing \emph{SpaceGAN}, we seek to further the agenda of spatial representations in deep learning. As many real-world applications of deep learning algorithms deal with geospatial data, tools tailored to these tasks are required \citep{Reichstein2019}. Nevertheless, the potential applications of \emph{SpaceGAN} go beyond the geospatial and other $2$-dimensional data domains. While the use of neighbourhood structures as well as the Moran's I metric allow for the handling of data distributed in $n$-dimensional space, we seek to confirm the applicability in future studies. Further potentially fruitful research directions include experiments with different GAN architectures, e.g. Wasserstein loss functions, and application studies with sensitive spatial data, which could be obfuscated using \emph{SpaceGAN} without loosing desirable statistical properties. 

%% file: appendix.tex
\section*{APPENDIX}\label{APPENDIX}

\subsection*{A. Experimental Data}

Here we provide a more elaborate description of the datasets used for evaluating \emph{Experiment 1} and \emph{Experiment 2}. 

\textbf{Toy 1}: We create a synthetic dataset of $n=400$ observations. Following the notation $\mathbf{d} = (\mathbf{x}, y, \mathbf{c})$, we first set the spatial resolution, i.e. the spatial coordinates $\mathbf{c} = (c^{(1)},c^{(2)})$:
\begin{equation}
    \mathbf{C} = \begin{bmatrix} 
            \mathbf{c}_{1,1} = (2.5,2.5) & \mathbf{c}_{1,2} = (7.5,2.5) & ... & \mathbf{c}_{1,400} = (97.5,2.5) \\
            \mathbf{c}_{2,1} = (2.5,7.5) & ... & ... & ... \\
            ... & ... & ... & ... \\
            \mathbf{c}_{400,1} = (2.5,97.5) & ... & ... & \mathbf{c}_{400,400} =(97.5,97.5) 
        \end{bmatrix} 
\end{equation}
We then add an independent feature $x$ as a random draw from a Gaussian distribution with mean $\mu=1$ and standard deviation $\sigma=0$:
\begin{equation}
    x \sim Normal(\mu,\sigma)
\end{equation}
Now, we create the target variable $y$ as a function of spatial coordinates $\mathbf{c}$ and the random noise $\mathbf{x}$ as follows:
\begin{equation}
    y = \sin x + (c^{(1)} - c^{(2)})^{2}
\end{equation}
The table below provides the summary statistics of the such constructed synthetic dataset.
\begin{table}[!htbp] \centering 
  \caption{Summary statistics of the \emph{Toy 1} synthetic dataset.} 
  \label{} 
\scalebox{0.7}{
\begin{tabular}{@{\extracolsep{5pt}}lccccccc} 
\\[-1.8ex]\hline 
\hline \\[-1.8ex] 
Statistic & \multicolumn{1}{c}{N} & \multicolumn{1}{c}{Mean} & \multicolumn{1}{c}{St. Dev.} & \multicolumn{1}{c}{Min} & \multicolumn{1}{c}{Pctl(25)} & \multicolumn{1}{c}{Pctl(75)} & \multicolumn{1}{c}{Max} \\ 
\hline \\[-1.8ex] 
$c^{(1)}$ & 400 & 50.000 & 28.868 & 2.500 & 26.250 & 73.750 & 97.500 \\ 
$c^{(2)}$ & 400 & 50.000 & 28.868 & 2.500 & 26.250 & 73.750 & 97.500 \\ 
$y$ & 400 & 0.000 & 1.000 & $-$0.846 & $-$0.731 & 0.426 & 3.743 \\ 
$x$  & 400 & $-$0.008 & 0.960 & $-$2.993 & $-$0.641 & 0.638 & 2.702 \\ 
\hline \\[-1.8ex] 
\end{tabular}} 
\end{table} 

\textbf{Toy 2}: We create a synthetic dataset of $n=841$ observations. We again start by setting the spatial resolution, i.e. the spatial coordinates $\mathbf{c} = (c^{(1)}, c^{(2)})$:
\begin{equation}
    \mathbf{C} = \begin{bmatrix} 
            \mathbf{c}_{1,1} = (1.75,1.75) & \mathbf{c}_{1,2} = (5.25,1.75) & ... & \mathbf{c}_{1,841} = (99.75,1.75) \\
            \mathbf{c}_{2,1} = (1.75,5.25) & ... & ... & ... \\
            ... & ... & ... & ... \\
            \mathbf{c}_{841,1} = (1.75,99.75) & ... & ... & \mathbf{c}_{841,841} = (99.75,99.75) 
        \end{bmatrix} 
\end{equation}
We again add an independent variable $x$ as a random draw from a Gaussian distribution with mean $\mu=1$ and standard deviation $\sigma=0$:
\begin{equation}
    x \sim Normal(\mu,\sigma)
\end{equation}
Lastly, we create the target variable $y$ as a more complex function of spatial coordinates $\mathbf{c}$ and the random noise $x$ as follows:
\begin{equation}
    y = \sin (c^{(1)} + c^{(2)}) * 2\pi +  \lfloor z \rfloor * 0.1 c^{(1)}
\end{equation}
where $z \sim U(1.75, 99.75) * 0.01$. The table below provides again provides the summary statistics.
\begin{table}[!htbp] \centering 
  \caption{Summary statistics of the \emph{Toy 2} synthetic dataset.} 
  \label{} 
\scalebox{0.7}{
\begin{tabular}{@{\extracolsep{5pt}}lccccccc} 
\\[-1.8ex]\hline 
\hline \\[-1.8ex] 
Statistic & \multicolumn{1}{c}{N} & \multicolumn{1}{c}{Mean} & \multicolumn{1}{c}{St. Dev.} & \multicolumn{1}{c}{Min} & \multicolumn{1}{c}{Pctl(25)} & \multicolumn{1}{c}{Pctl(75)} & \multicolumn{1}{c}{Max} \\ 
\hline \\[-1.8ex] 
$c^{(1)}$ & 841 & 50.750 & 29.301 & 1.750 & 26.250 & 75.250 & 99.750 \\ 
$c^{(2)}$ & 841 & 50.750 & 29.301 & 1.750 & 26.250 & 75.250 & 99.750 \\ 
$y$ & 841 & $-$0.000 & 1.000 & $-$2.294 & $-$0.606 & 0.631 & 2.488 \\ 
$x$ & 841 & $-$0.032 & 1.021 & $-$3.372 & $-$0.700 & 0.660 & 3.495 \\ 
\hline \\[-1.8ex] 
\end{tabular}}
\end{table} 

\textbf{California Housing}: This real world dataset, introduced by \citep{KelleyPace2003}, is widely popular for analyzing spatial patterns and accessible via \emph{Kaggle}\footnote{See:\url{https://www.kaggle.com/camnugent/california-housing-prices}} (it is also integrated into \emph{sklearn}\footnote{See:\url{https://scikit-learn.org/stable/modules/generated/sklearn.datasets.fetch_california_housing.html}}). The table below provides an overview of the features and their statistical properties:
\begin{table}[!htbp] \centering 
  \caption{Summary statistics of the \emph{California Housing} dataset.} 
  \label{}
\scalebox{0.7}{
\begin{tabular}{@{\extracolsep{5pt}}lccccccc} 
\\[-1.8ex]\hline 
\hline \\[-1.8ex] 
Statistic & \multicolumn{1}{c}{N} & \multicolumn{1}{c}{Mean} & \multicolumn{1}{c}{St. Dev.} & \multicolumn{1}{c}{Min} & \multicolumn{1}{c}{Pctl(25)} & \multicolumn{1}{c}{Pctl(75)} & \multicolumn{1}{c}{Max} \\ 
\hline \\[-1.8ex] 
longitude & 20,640 & $-$119.570 & 2.004 & $-$124.350 & $-$121.800 & $-$118.010 & $-$114.310 \\ 
latitude & 20,640 & 35.632 & 2.136 & 32.540 & 33.930 & 37.710 & 41.950 \\ 
housing\_median\_age & 20,640 & 28.639 & 12.586 & 1 & 18 & 37 & 52 \\ 
total\_rooms & 20,640 & 2,635.763 & 2,181.615 & 2 & 1,447.8 & 3,148 & 39,320 \\ 
total\_bedrooms & 20,433 & 537.871 & 421.385 & 1.000 & 296.000 & 647.000 & 6,445.000 \\ 
population & 20,640 & 1,425.477 & 1,132.462 & 3 & 787 & 1,725 & 35,682 \\ 
households & 20,640 & 499.540 & 382.330 & 1 & 280 & 605 & 6,082 \\ 
median\_income & 20,640 & 3.871 & 1.900 & 0.500 & 2.563 & 4.743 & 15.000 \\ 
median\_house\_value & 20,640 & 206,855.800 & 115,395.600 & 14,999 & 119,600 & 264,725 & 500,001 \\ 
\hline \\[-1.8ex] 
\end{tabular}}
\end{table}
We can break the dataset down into the familiar notation $\mathbf{d} = (\mathbf{x}, y,\mathbf{c})$ as follows:
\begin{equation}
    \mathbf{c} = (longitude,latitude) 
\end{equation}
\begin{equation}
\begin{gathered}
    \mathbf{x} = (housing\_median\_age,total\_rooms,total\_bedrooms,population, \\ households,median\_income)
\end{gathered}
\end{equation}
\begin{equation}
    y = (median\_house\_value) 
\end{equation}

\subsection*{B. Experimental Setting}

The tables below provide details on architecture and configuration of the neural networks used in \emph{SpaceGAN} during our experiments. Note that the kernel size parameter for \textbf{Toy 1} and \textbf{Toy 2} corresponds to the queen neighbourhood (for discrete spatial data) outlined in \ref{fig1} and is the same neighbourhood that is used for spatial conditioning and spatial cross validation (see Appendix E). The kernel size for \textbf{California Housing 15} and \textbf{California Housing 50} corresponds to same kNN-neighbourhood (with $k=50$) that is used for spatial conditioning and spatial cross-validation.

\begin{table}[h!]
	\centering
	\tiny
	\caption{Dataset-specific configurations of the \emph{SpaceGAN} architecture for the experiments.} \label{spacegan-configs1}
	\begin{tabular}{cc}
		\hline
		\hline
		\multicolumn{1}{c}{\textbf{Parameter}} & \multicolumn{1}{c}{\textbf{Values}} \\ \hline
		Architecture & 1D-CNN \\
		Number of hidden layers & 1 \\
		Training steps & 20000 \\
		Batch Size & 100 \\
		Optimizer & Stochastic Gradient Descent \\
		Optimizer Parameters & learning rate = 0.01 \\
		Noise prior $p_{\mathbf{z}}(\mathbf{z})$ & $N(0, 1)$ \\
		Snapshot frequency ($snap$) & 500 \\
		Number of samples for evaluation & $C =$ 500 \\
		Input features scaling function & $Z$-score (standardization) \\
		Target scaling function & $Z$-score (standardization) \\
		\hline
		\hline
	\end{tabular}
\end{table}

\begin{table}[h!]
	\centering
	\tiny
	\caption{Overview of the general \emph{SpaceGAN} architecture and its hyperparameters.} \label{spacegan-configs2}
	\begin{tabular}{c|cccc}
		\hline
		\hline
		\textbf{Parameter} & \textbf{Toy 1} & \textbf{Toy 2} & \textbf{California 15} & \textbf{California 50}  \\
		\hline
		($G$, $D$) filters ($|\mathcal{N}_{i}|$) & (50, 50) & (100, 100) & (100, 100) & (200, 200) \\
		($G$, $D$) kernel size & (8, 8) & (8, 8) & (15, 15) & (50, 50)  \\
		($G$, $D$) hidden layer function & (relu, tanh) & (relu, tanh) & (relu, tanh) & (relu, tanh) \\
		($G$, $D$) output layer function & (linear, sigmoid) & (linear, sigmoid) & (linear, sigmoid) & (linear, sigmoid) \\
		Noise dimension $dim(\mathbf{z})$ & 8 & 8 & 15 & 15\\
		\hline
		\hline
	\end{tabular}
\end{table}

\subsection*{D. Training convergence: $MIE$ vs. $RMSE$}

\begin{figure}[!htbp]
\centering
\begin{subfigure}{.5\textwidth}
  \centering
  \includegraphics[width=.9\linewidth]{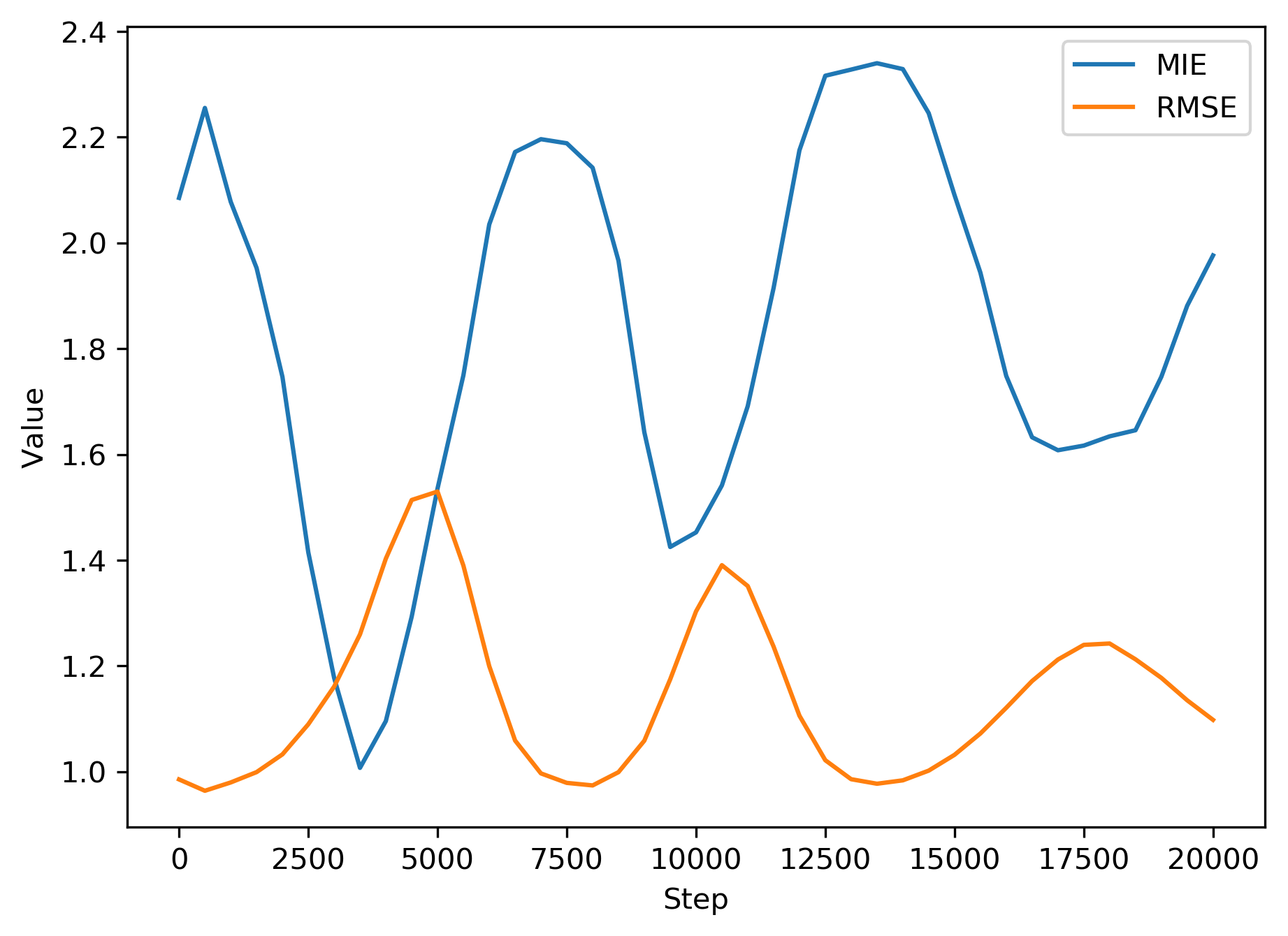}
  \caption{Toy 1}
  \label{fig:sub1}
\end{subfigure}%
\begin{subfigure}{.5\textwidth}
  \centering
  \includegraphics[width=.9\linewidth]{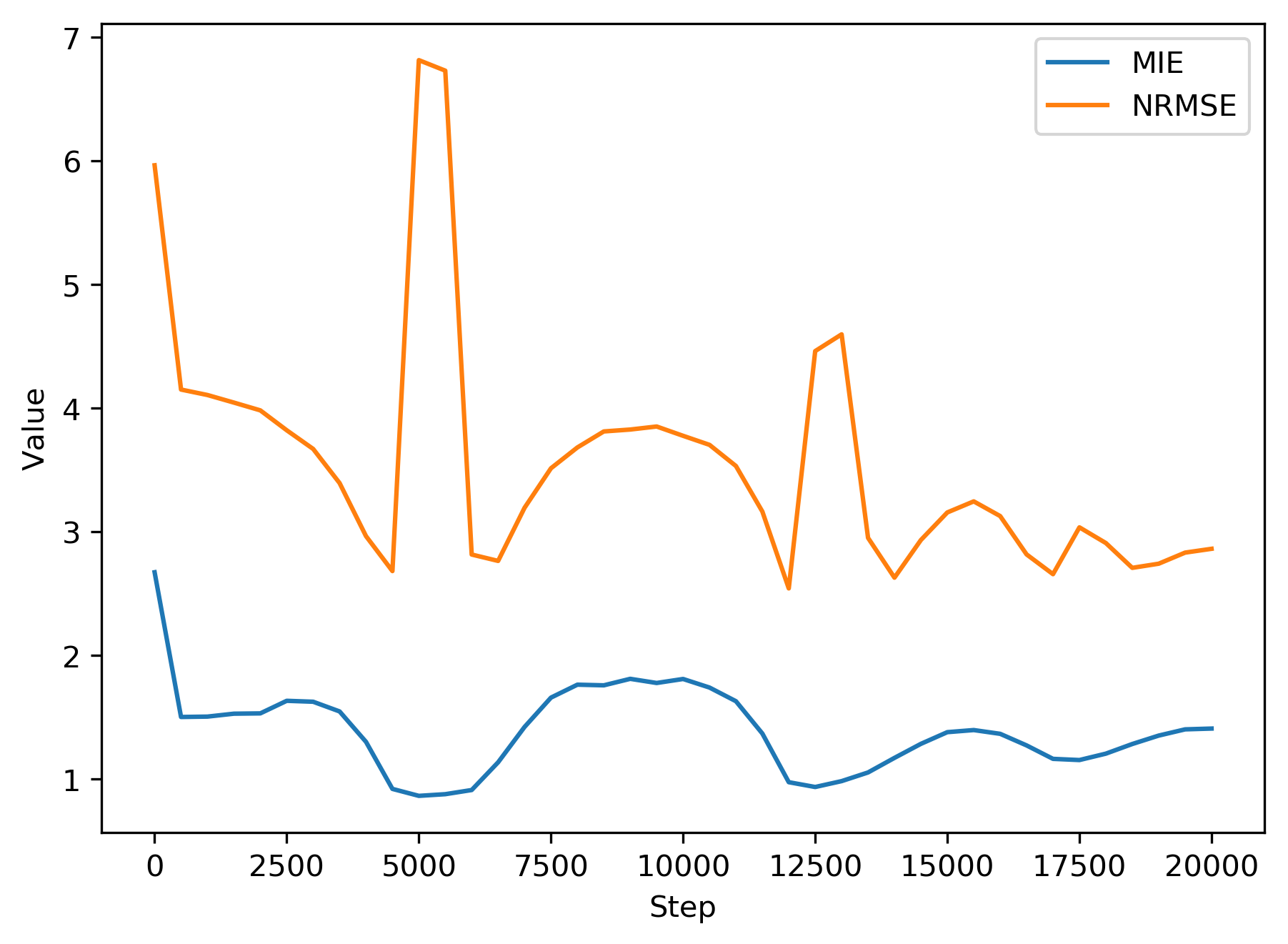}
  \caption{California Housing 50}
  \label{fig:sub2}
\end{subfigure}
\caption{MIE and RMSE evolution through a typical SpaceGAN training cycle.} \label{mie-rmse-curves-spacegan}
\end{figure}

During \textit{Experiment 2}, we compare the convergence (and performance) of two \emph{SpaceGAN} implementations, one using $MIE$ and one using $RMSE$ as convergence criterion. For completeness, we define the $RMSE$ (root mean squared error) as follows:
\begin{equation}
    RMSE = \sqrt{ \sum^{n}_{i=1} (y_{i} - \hat{y_{i}})^{2}}
\end{equation}
The figure below shows \emph{SpaceGAN} training using the different convergence criteria for \textbf{Toy 1} and \textbf{California Housing 50} over training steps $tsteps$, during a typical training cycle. Interestingly, for \textbf{Toy 1}, both criteria are almost antithetic, that is a local minimum for $MIE$ convergence approximately relates to a local maximum for $RMSE$ convergence in the same training step. Moreover, $RMSE$ struggles to provide assistance for when a convergence point is reached, as it shows several local minima of approximately similar value. This point is also true for the \textbf{California Housing 50} dataset. The $MIE$ criterion however appears to have a relatively stable minimum at the first local minimum point. 

\subsection*{E. Spatial Cross-Validation}

\begin{wrapfigure}{r}{0.4\textwidth}
\vskip -0.1in
\centering
\includegraphics[width=0.4\textwidth]{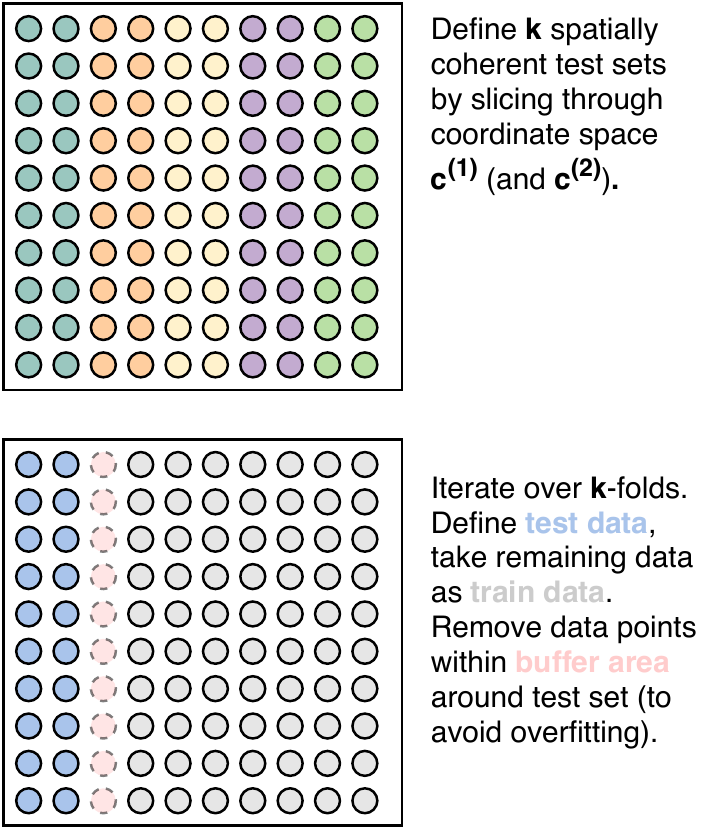}
%\vskip -0.1in
\caption{Illustration of the spatial $k$-fold cross validation process.} \label{supp1}
%\vskip -0.15in
%\vskip -0.5in
\end{wrapfigure}
We use a variation of $k$-fold spatial cross-validation \citep{Pohjankukka2017} to evaluate all our experiments. The goal of spatial cross-validation is to check for generalizability of spatial models and to avoid overfitting. In a naive cross-validation setting with spatial data, this can occur when training and test points are spatially to close. Assuming some spatial dependency between nearby points, this would roughly relate to training on the test set. Hence, we need to create a so-called buffer area around the test set within which we remove all data points from the training set. Assuming a set of data points $\mathcal{D} = (\mathbf{d}_{1}, \mathbf{d}_{2},..., \mathbf{d}_{n})$, we first create $k$ spatially coherent test sets. In our case, we do this by slicing through each of the two dimensions of the coordinate space $\mathbf{c}$ five times with equal binning, thus creating $10$ folds of the same width. This leaves us with a set of $10$ test sets $\mathcal{D}_{test}= (\mathcal{D}_{test}^{(1)},\mathcal{D}_{test}^{(2)},...,\mathcal{D}_{test}^{(10)})$. We now define the training set  $\mathcal{D}_{train}^{(k)}$ as all points in set $\mathcal{D}$ which are not part of the test set $\mathcal{D}_{test}^{(k)}$ and which are not neighbouring points of the test set points, thus creating a buffer area: $\mathcal{D}_{train}^{(k)} = \mathcal{D} \notin \mathcal{D}_{test}^{(k)}, \notin \mathcal{N}_{\mathcal{D}_{test}^{(k)}}$. As a quick example, for the \textbf{California Housing 50} dataset, we would define the test set, then exclude all point which are not part of the test set, but are one of the $50$-nearest-neighbours of one of the test set points. The remaining, not excluded points provide the training set. While we chose to define the buffer zone according to the neighbourhood based spatial weights matrix $w$, other methods such as defining a deadzone area using a radius around the test set are also applicable. The spatial $k$-folds cross validation process is outlined in Figure \ref{supp1} to the right. 

\subsection*{F. Experimental Results}

Here, we want to provide some higher-resolution images of the \emph{SpaceGAN}-augmented data across the three example datasets. Please note again that all synthetic samples are based on out-of-sample extrapolations from the respective generator (\emph{SpaceGAN} or GP). 

\newpage

\begin{sidewaysfigure}[!ht]
    \begin{centering}
        \includegraphics[width=\linewidth]{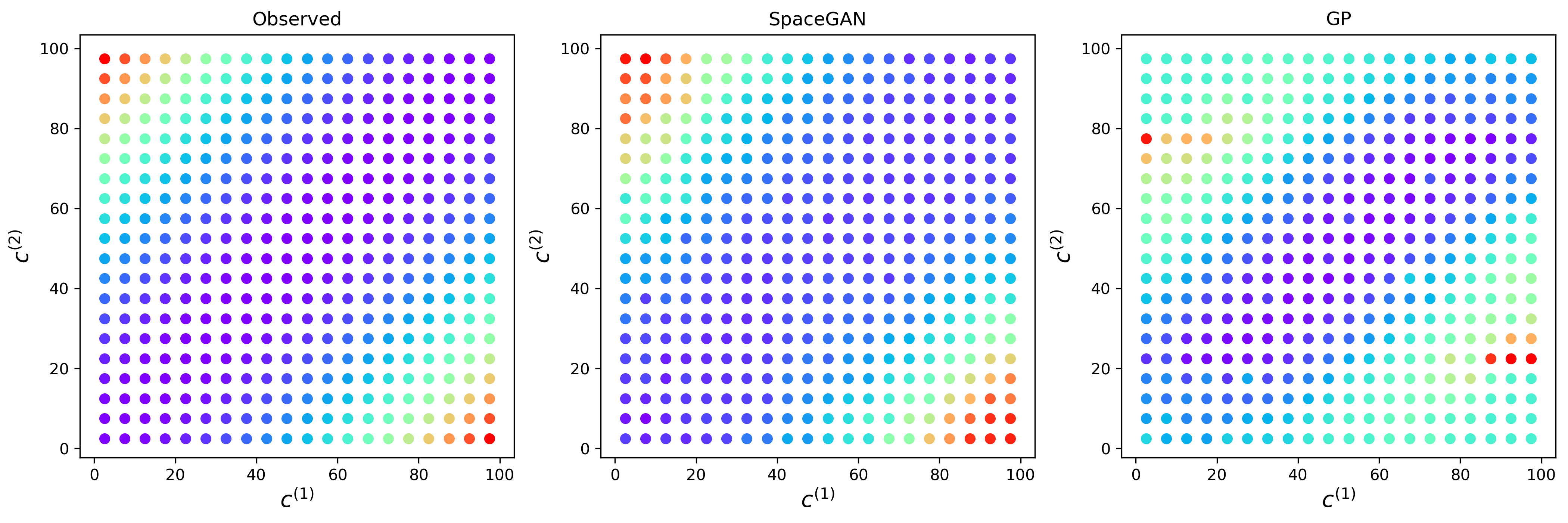}
        \includegraphics[width=\linewidth]{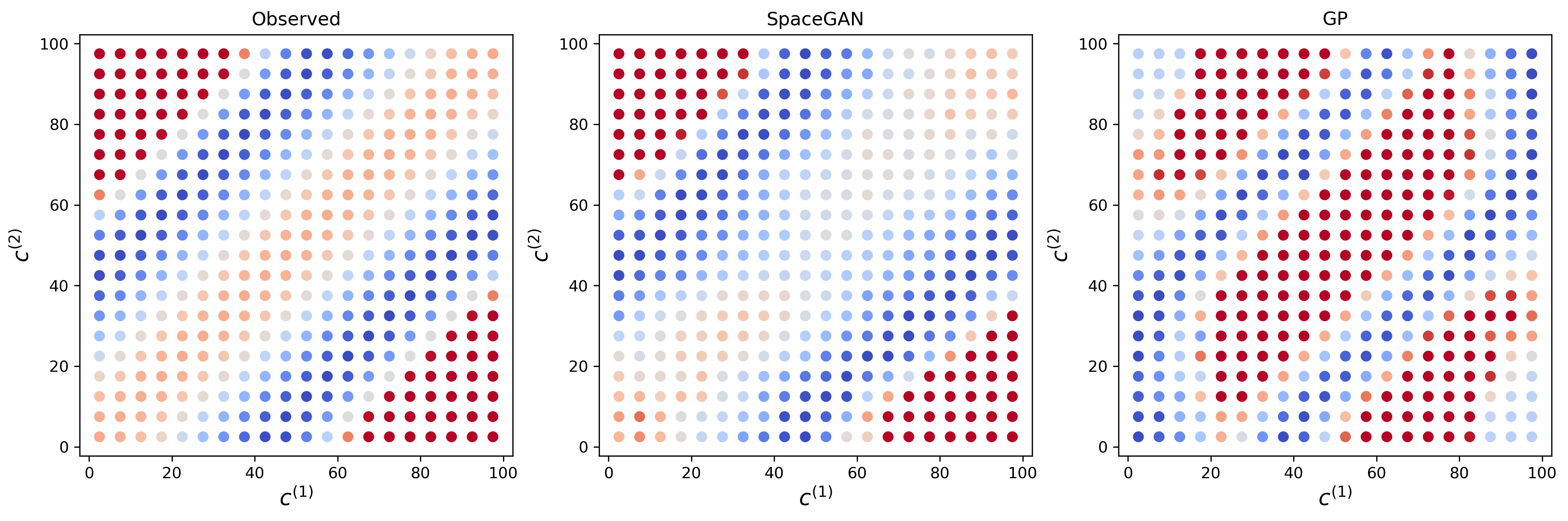}
        \caption{Real vs. synthetic data for the \textbf{Toy 1} dataset. \emph{SpaceGAN} and GPs are used for data augmentation. The upper row shows the target vector $\mathbf{y}$, the lower row it's local spatial autocorrelation $I(\mathbf{y})$. All synthetic data is generated through out-of-sample extrapolation with spatial cross-validation.} \label{supp2}
    \end{centering}
\end{sidewaysfigure}

\newpage

\begin{sidewaysfigure}[!ht]
    \begin{centering}
        \includegraphics[width=\linewidth]{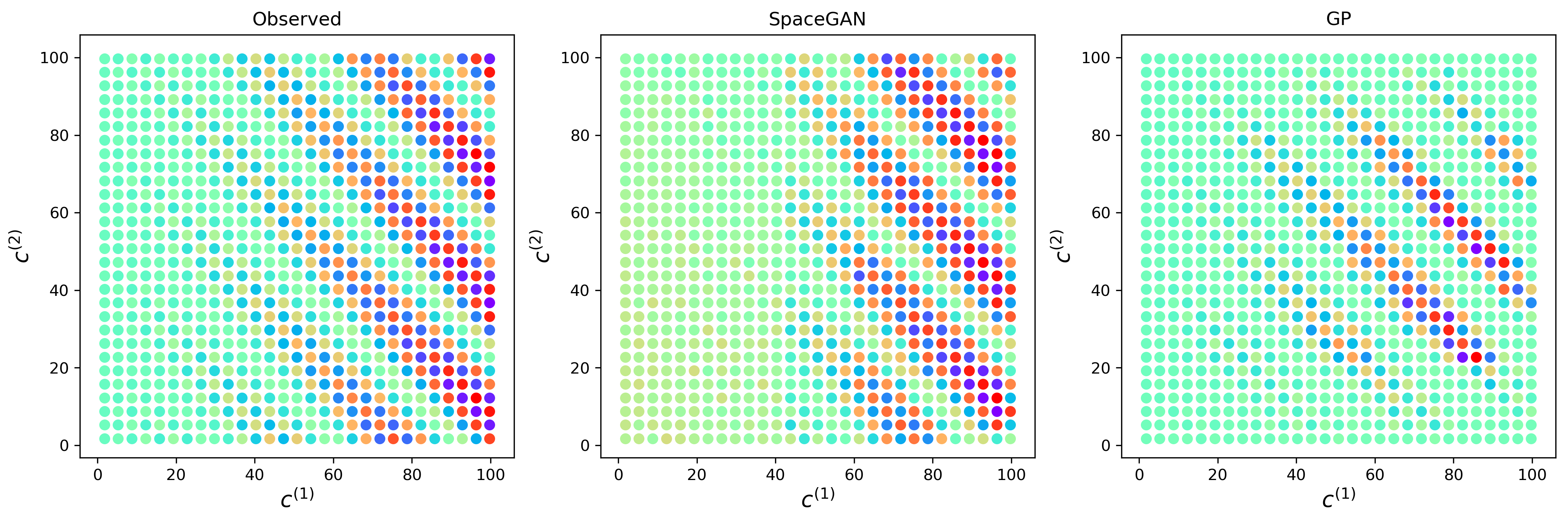}
        \includegraphics[width=\linewidth]{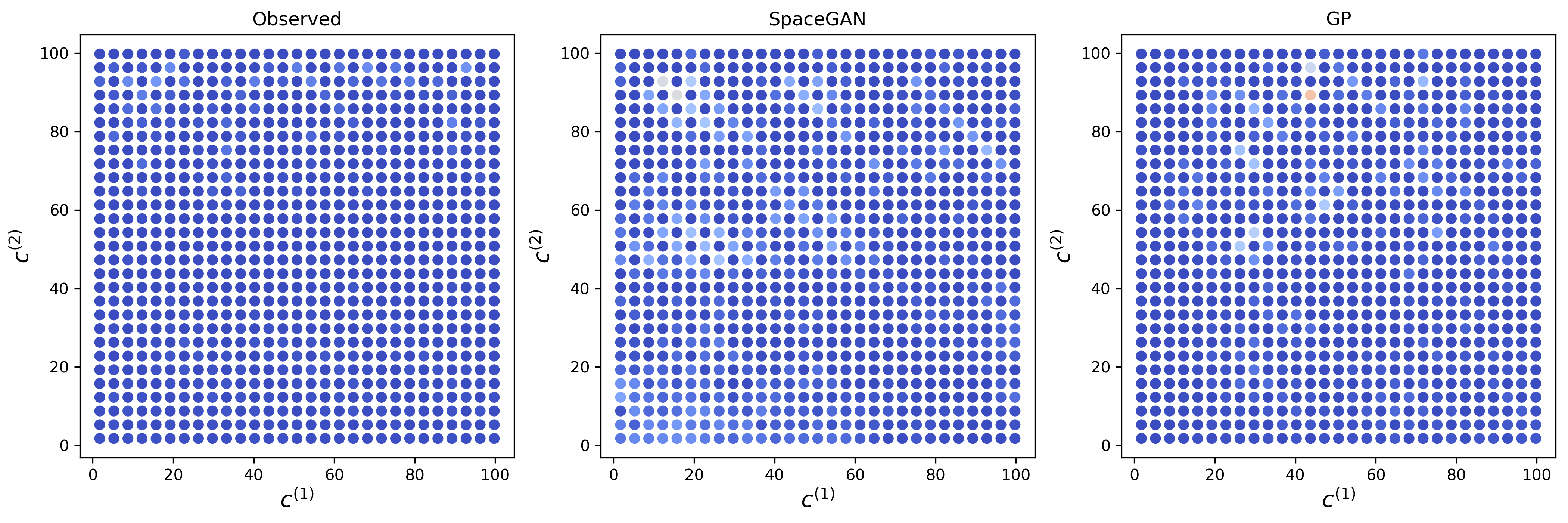}
        \caption{Real vs. synthetic data for the \textbf{Toy 2} dataset. \emph{SpaceGAN} and GPs are used for data augmentation. The upper row shows the target vector $\mathbf{y}$, the lower row it's local spatial autocorrelation $I(\mathbf{y})$. All synthetic data is generated through out-of-sample extrapolation with spatial cross-validation.} \label{supp3}
    \end{centering}
\end{sidewaysfigure}

\newpage

\begin{sidewaysfigure}[!ht]
    \begin{centering}
        \includegraphics[width=\linewidth]{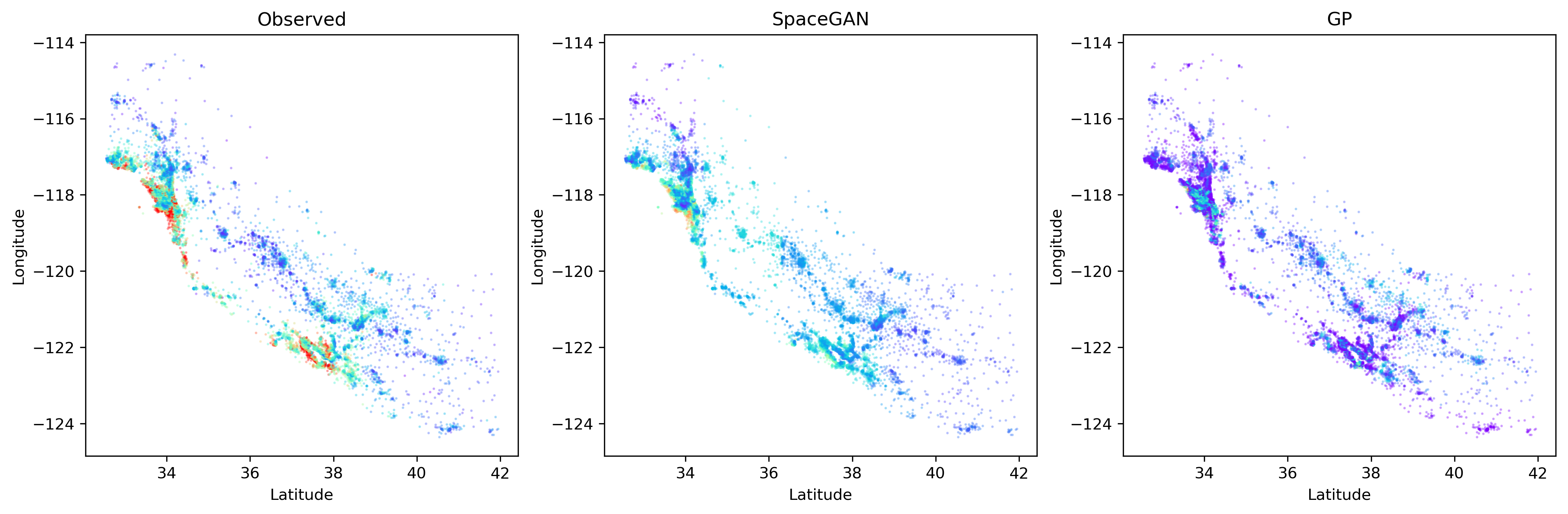}
        \includegraphics[width=\linewidth]{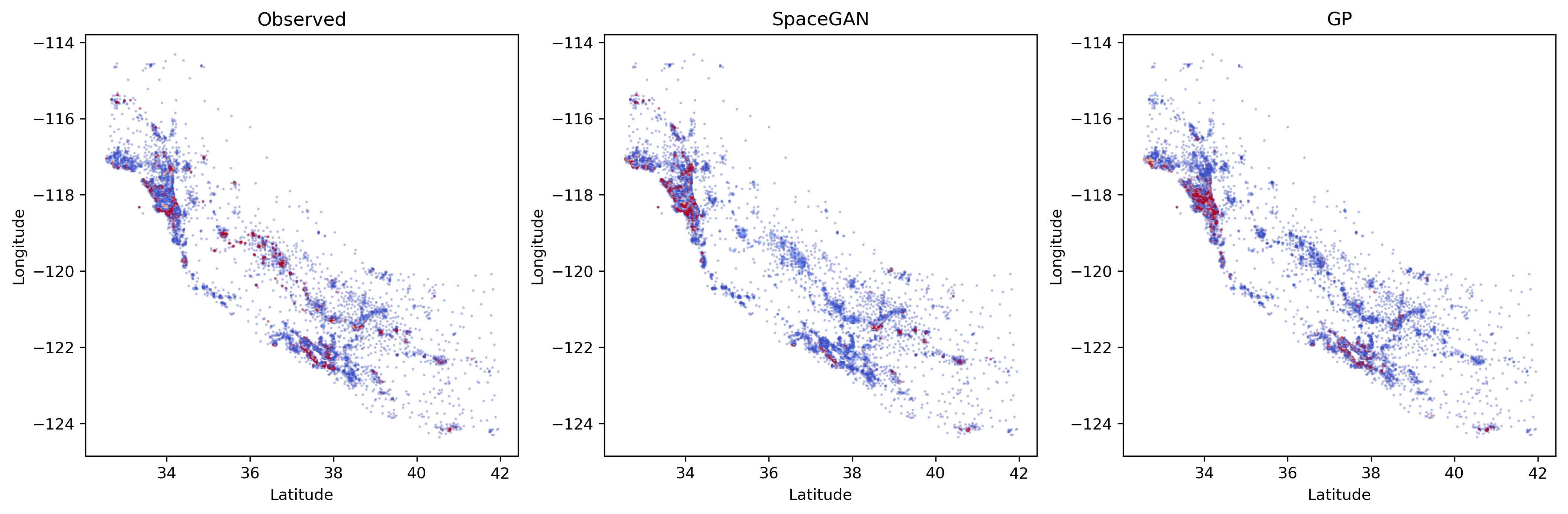}
        \caption{Real vs. synthetic data for the \textbf{California Housing 15} dataset. Here, we use the 15 nearest neighbours of each datapoint for (1) defining the ConvNet kernel size in $G,D$, (2) calculating $MIE$ and (3) spatial cross-validation. \emph{SpaceGAN} and GPs are used for data augmentation. The upper row shows the target vector $\mathbf{y}$, the lower row it's local spatial autocorrelation $I(\mathbf{y})$. All synthetic data is generated through out-of-sample extrapolation with spatial cross-validation.} \label{supp4}
    \end{centering}
\end{sidewaysfigure}

\newpage

\begin{sidewaysfigure}[!ht]
    \begin{centering}
        \includegraphics[width=\linewidth]{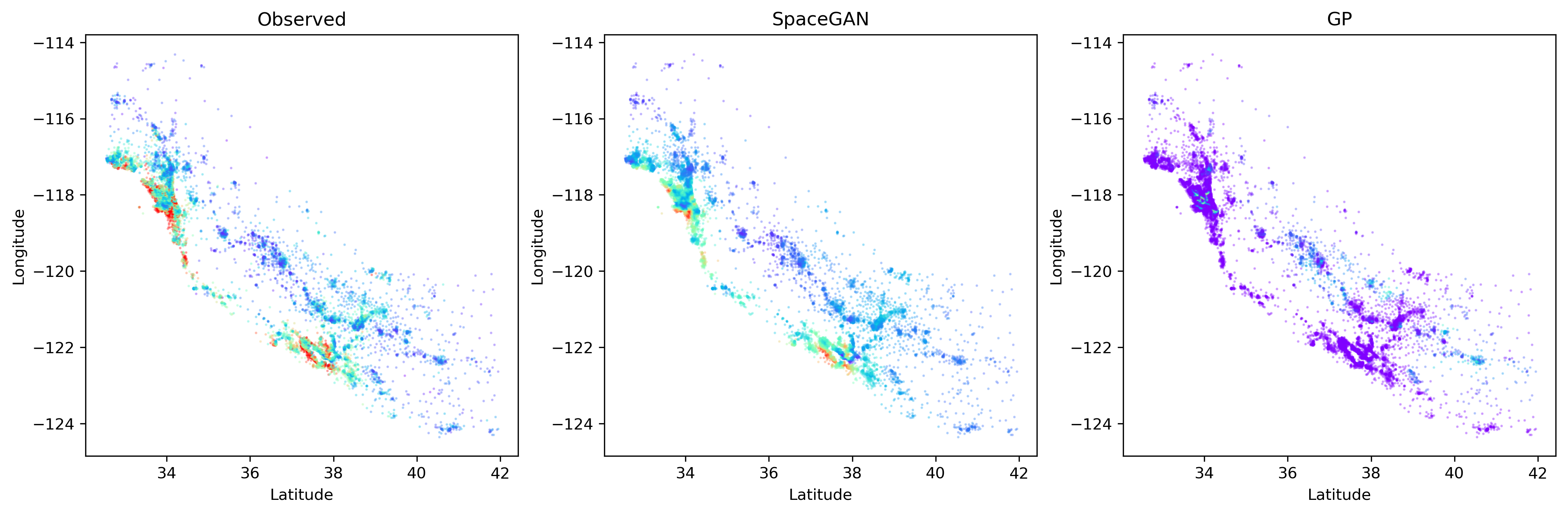}
        \includegraphics[width=\linewidth]{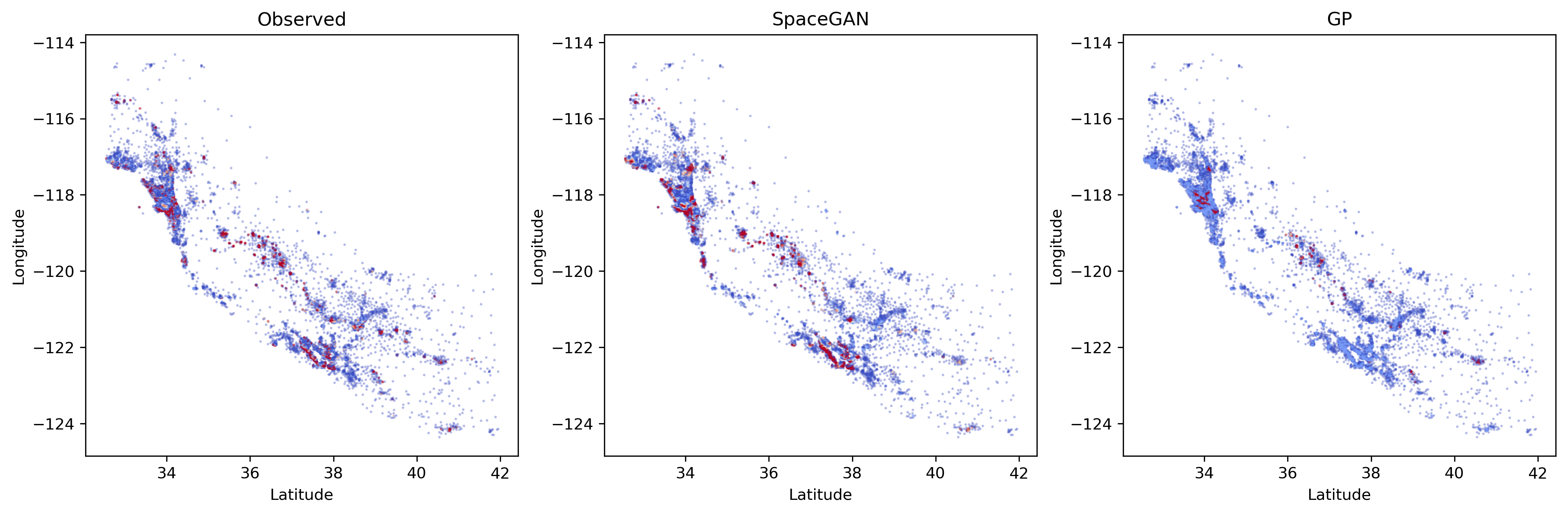}
        \caption{Real vs. synthetic data for the \textbf{California Housing 50} dataset. Here, we use the 50 nearest neighbours of each datapoint for (1) defining the ConvNet kernel size in $G,D$, (2) calculating $MIE$ and (3) spatial cross-validation. \emph{SpaceGAN} and GPs are used for data augmentation. The upper row shows the target vector $\mathbf{y}$, the lower row it's local spatial autocorrelation $I(\mathbf{y})$. All synthetic data is generated through out-of-sample extrapolation with spatial cross-validation.} \label{supp5}
    \end{centering}
\end{sidewaysfigure}